\documentclass[11pt,draftclsnofoot,onecolumn]{IEEEtran}

\usepackage{graphicx,slashbox}
\usepackage{epsfig}
\usepackage{amsmath}
\usepackage{amssymb}
\usepackage{subfigure}
\usepackage{wrapfig}
\usepackage{times,color}
\usepackage{cite,footnote,xspace,syntonly,algorithm,algorithmic,bm}
\input{mysymbol.sty}
\long\def\symbolfootnote[#1]#2{\begingroup
\def\thefootnote{\fnsymbol{footnote}}
\footnote[#1]{#2}\endgroup} \psfull

\begin{document}
\title{\huge Robust Nonparametric Regression via Sparsity Control\\
with Application to Load Curve Data Cleansing$^\dag$}

\author{{\it Gonzalo~Mateos and Georgios~B.~Giannakis~(contact author)$^\ast$}}

\markboth{IEEE TRANSACTIONS ON SIGNAL PROCESSING (SUBMITTED)}
\maketitle \maketitle \symbolfootnote[0]{$\dag$ Work in this
paper was supported by the NSF grants CCF-0830480, 1016605 and
ECCS-0824007, 1002180. Part of the paper will appear in the
\emph{Intl. Conf. on Acoust., Speech, and Signal Proc.}, Prague, Czech 
Republic, May. 22-27,
2011.} \symbolfootnote[0]{$\ast$ The authors are with the Dept.
of Electrical and Computer Engineering, University of
Minnesota, 200 Union Street SE, Minneapolis, MN 55455. Tel/fax:
(612)626-7781/625-4583; Emails:
\texttt{\{mate0058,georgios\}@umn.edu}}

\vspace*{-80pt}
\begin{center}
\small{\bf Submitted: }\today\\
\end{center}
\vspace*{10pt}

\thispagestyle{empty}\addtocounter{page}{-1}
\begin{abstract}
Nonparametric methods are widely applicable to statistical
inference problems, since they rely on a few modeling
assumptions. In this context, the fresh look advocated here
permeates benefits from variable selection and compressive
sampling, to robustify nonparametric regression against
outliers -- that is, data markedly deviating from
the  postulated models. 
A variational counterpart to least-trimmed
squares regression is shown closely related to an
$\ell_0$-(pseudo)norm-regularized estimator, that encourages
\textit{sparsity} in a vector explicitly modeling the outliers.
This connection suggests efficient solvers based
on convex relaxation, which lead naturally to a variational
M-type estimator equivalent to the least-absolute shrinkage and selection 
operator (Lasso). Outliers are identified
by judiciously tuning regularization parameters, which amounts
to controlling the sparsity of the outlier vector along the
whole \textit{robustification} path of Lasso solutions. 
Reduced bias and enhanced generalization capability 
are attractive features of an improved estimator obtained after
replacing the $\ell_0$-(pseudo)norm with a nonconvex surrogate. The novel 
robust 
spline-based smoother is adopted to
cleanse \textit{load curve} data, a key task
aiding operational decisions in the envisioned smart grid system.
Computer simulations and tests on real load curve data 
corroborate the effectiveness of the novel sparsity-controlling robust
estimators.
\end{abstract}
\vspace*{-5pt}
\begin{center} \bfseries EDICS Category: SSP-NPAR, MLR-LEAR. \end{center}
%
\newpage

\section{Introduction}\label{sec:intro}

Consider the classical problem of function estimation, in which
an input vector
$\mathbf{x}:=[x_1,\ldots,x_p]^\prime\in\mathbb{R}^p$ is given,
and the goal is to predict the real-valued scalar response
$y=f(\mathbf{x})$. Function $f$ is unknown, to be estimated
from a training data set
$\mathcal{T}:=\{y_i,\mathbf{x}_i\}_{i=1}^{N}$. When $f$ is
assumed to be a member of a finitely-parameterized family of
functions, standard (non-)linear regression techniques can
be adopted. If on the other hand, one is only willing to assume
that $f$ belongs to a (possibly infinite dimensional) space of
``smooth'' functions $\mathcal{H}$, then a \textit{nonparametric}
approach is in order, and this will be the focus of this work. 

Without further constraints beyond $f\in\mathcal{H}$,
functional estimation from finite data is an ill-posed problem.
To bypass this challenge, the problem is typically solved by
minimizing appropriately regularized criteria, allowing one to
control model complexity; see,
e.g.,~\cite{regnet_svm,Tikhonov}. It is then further assumed that
$\mathcal{H}$ has the structure of a reproducing kernel Hilbert
space (RKHS), with corresponding positive definite reproducing
kernel function $K(\cdot,\cdot):\mathbb{R}^p\times
\mathbb{R}^p\rightarrow\mathbb{R}$, and norm denoted by
$\|\cdot\|_{\mathcal{H}}$. Under the formalism of 
\textit{regularization networks}, one seeks $\hat f$ as the solution to the 
variational problem
\begin{equation}\label{eq:regnets}
\min_{f\in\mathcal{H}}\left[\sum_{i=1}^N 
V(y_i-f(\mathbf{x}))+\mu\|f\|_{\mathcal{H}}^2\right]
\end{equation}
where $V(\cdot)$ is a convex loss function, and $\mu\geq 0$ controls complexity
 by weighting the effect of the smoothness functional $\|f\|_{\mathcal{H}}^2$. 
Interestingly, the Representer Theorem asserts that the unique solution of 
\eqref{eq:regnets} is finitely parametrized and has the form 
$\hat{f}(\mathbf{x})=\sum_{i=1}^{N}\beta_i K(\mathbf{x},\mathbf{x}_i)$,
where $\{\beta_i\}_{i=1}^N$ can be obtained from $\mathcal{T}$; see e.g., 
~\cite{po_gi_89,wahba}. 
 Further details on 
RKHS, and in particular on the evaluation of
$\|f\|_{\mathcal{H}}$, can be found in e.g.,~\cite[Ch.
1]{wahba}. A fundamental relationship between model complexity control 
and generalization capability, i.e., the predictive ability
of $\hat f$ beyond the training set, 
was formalized in~\cite{Vapnik_Learning_Theory}.

The generalization error performance of approaches that minimize the sum
of squared model residuals [that is $V(u)=u^2$ in 
\eqref{eq:regnets}] regularized by a term of the form
$\|f\|_{\mathcal{H}}^2$, is 
degraded in the presence
of outliers. This is because the least-squares (LS) part
of the cost is not robust, and can result in severe overfitting of
the (contaminanted) training data~\cite{huber}. 
Recent efforts have
considered replacing the squared loss with a robust counterpart
such as Huber's function, or its variants, but lack a
data-driven means of selecting the proper threshold that
determines which datum is considered an
outlier~\cite{robust_kernel_regression}; see 
also~\cite{robust_svr}. Other approaches have 
instead relied 
on the so-termed
$\epsilon$-insensitive loss function, originally proposed to solve function 
approximation problems using support vector machines 
(SVMs)~\cite{Vapnik_Learning_Theory}. These family of estimators often referred
 to as support vector regression (SVR), have been shown to enjoy 
 robustness properties; see e.g.,~\cite{SVR_tutorial,girosi97,SSVR} and 
references 
therein. 
In~\cite{RSVR}, improved performance in the presence of outliers is achieved 
by refining the SVR solution through a subsequent robust learning phase. 

The starting point here is a variational least-trimmed squares
(VLTS) estimator, suitable for robust function approximation in
$\mathcal{H}$ (Section \ref{sec:model}). It is established that VLTS is closely
 related to an
(NP-hard) $\ell_0$-(pseudo)norm-regularized estimator, adopted
to fit a regression model that explicitly incorporates an
unknown \textit{sparse} vector of outliers~\cite{fuchs}. As in
compressive sampling (CS)~\cite{tropp_relax}, efficient
(approximate) solvers are obtained in Section \ref{sec:spacor} by replacing the
 outlier
vector's $\ell_0$-norm with its closest convex
approximant, the $\ell_1$-norm. This leads naturally to a
variational M-type estimator of $f$, also shown equivalent to a
least-absolute shrinkage and selection operator
(Lasso)~\cite{tibshirani_lasso} on the vector of outliers (Section 
\ref{ssec:solve_l1unconstr}). A
tunable parameter in Lasso \textit{controls} the \textit{sparsity} of the
estimated vector, and the number of outliers as a byproduct.
Hence, effective methods to select this parameter are of
paramount importance.

The link between $\ell_1$-norm regularization and robustness
was also exploited for parameter (but not function) estimation
in~\cite{fuchs} and~\cite{rao_robust}; see 
also~\cite{Wright2010-IT} for related ideas 
in the context of face recognition, and error correction 
codes~\cite{candes_error_correction}. In~\cite{fuchs} however,
the selection of Lasso's tuning parameter is only justified for
Gaussian training data; whereas a fixed value motivated by CS
error bounds is adopted in the Bayesian formulation
of~\cite{rao_robust}. Here instead, a more general and
systematic approach is pursued in Section \ref{ssec:params}, building on 
contemporary
algorithms that can efficiently compute all
\textit{robustifaction} paths of Lasso solutions, i.e., for all
values of the tuning
parameter~\cite{Efron_lars_2004,friedman_2009,Wright_tsp_2009}.
In this sense, the method here capitalizes on but \textit{is
not limited to} sparse settings, since one can examine all
possible sparsity levels along the robustification path. An
estimator with reduced bias and improved generalization
capability is obtained in Section \ref{sec:nonconvex}, after replacing the
$\ell_0$-norm with a nonconvex surrogate, instead of
the $\ell_1$-norm that introduces
bias~\cite{tibshirani_lasso,zou_adalasso}. Simulated tests
demonstrate the effectiveness of the novel approaches in 
robustifying thin-plate smoothing splines~\cite{duchon} (Section 
\ref{ssec:robust_splines}), and in estimating the $\textrm{sinc}$ function 
(Section \ref{ssec:sinc_approx}) -- a paradigm typically adopted to assess 
performance of robust function approximation
approaches~\cite{RSVR,robust_kernel_regression}.

\begin{figure}[h]
\begin{center}
\includegraphics[width=0.7\linewidth]{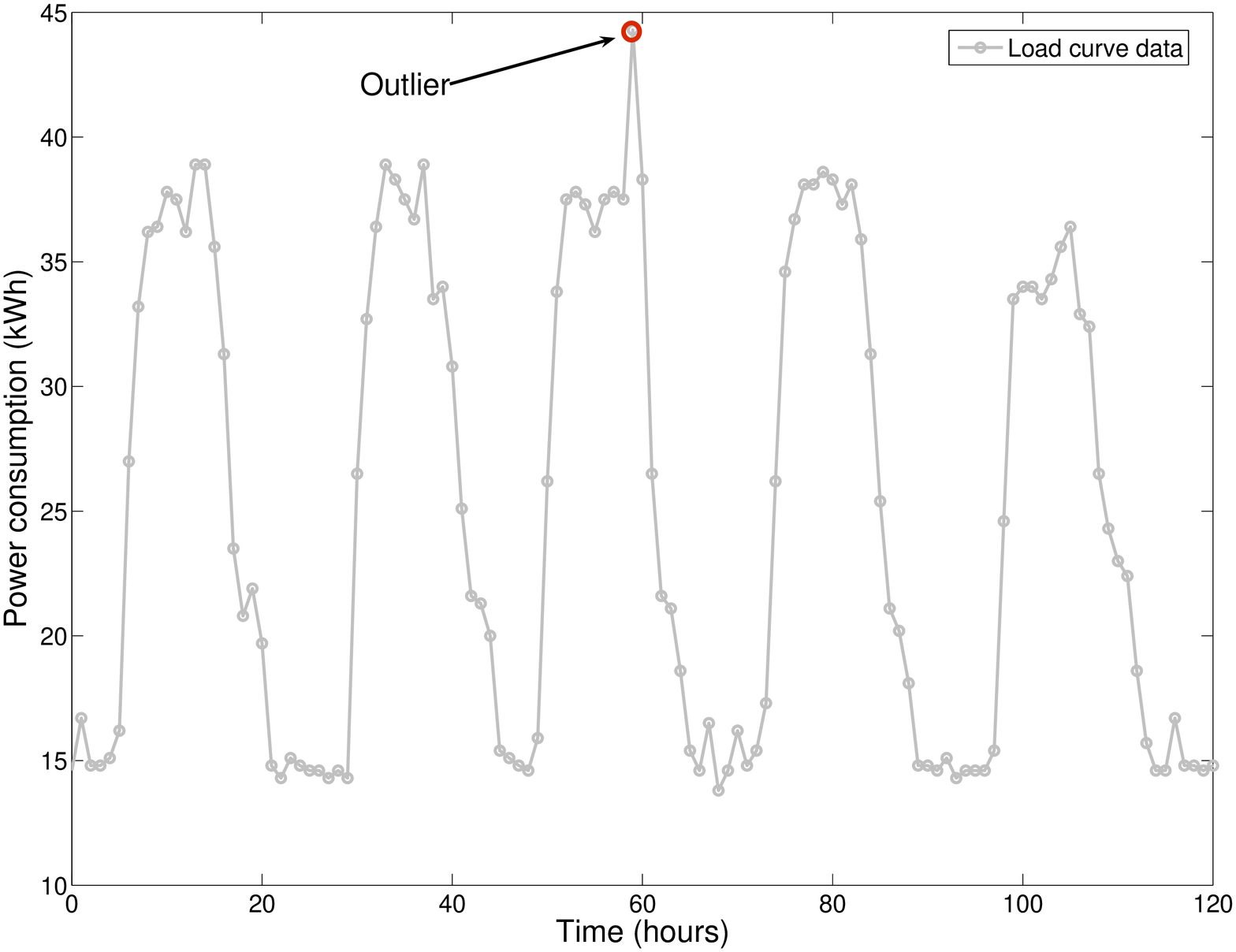}\\
\vspace{-0.5cm}
\caption{Example of load curve data with outliers.}\label{fig:Fig_0}
\end{center}
\end{figure}

The motivating application behind the robust nonparametric 
methods of this paper is \textit{load curve 
cleansing}~\cite{load_curve_cleansing} -- a critical task in power systems 
engineering and management. Load curve
data (also known as load profiles) refers to the electric energy 
consumption periodically recorded
by meters at specific points across the power grid, e.g., end user-points and 
substations. Accurate load profiles are critical assets
aiding operational decisions in the envisioned smart grid 
system~\cite{smartgrid2}; see 
also~\cite{smartgrid1,smartgrid3,load_curve_cleansing}.
However, in the process of acquiring and transmitting such
massive volumes of information to a central processing unit, data is often 
noisy, corrupted, or
lost altogether. This could be due to several reasons including meter 
misscalibration or outright failure, as well as 
communication errors due to noise, network congestion, and connectivity 
outages; see Fig. 
\ref{fig:Fig_0} for an example. In
addition, data significantly deviating from nominal load models (outliers) are 
not 
uncommon, and could be attributed to unscheduled maintenance leading to 
shutdown of
heavy industrial loads, weather constraints, holidays, strikes, and major 
sporting events, just to name a few.

In this context,
it is critical to effectively reject outliers, and replace the 
contaminated data with `healthy' load predictions, i.e., to cleanse the load 
data. While most utilities carry out this task manually based on their own 
personnel's know-how, a first scalable and principled approach to load profile 
cleansing 
which is based on statistical learning methods was 
recently proposed in~\cite{load_curve_cleansing}; which also includes an 
extensive literature review on the related problem of outlier identification in
 time-series. After estimating the regression function $f$ via either B-spline 
or Kernel smoothing, pointwise confidence intervals are constructed based on 
$\hat f$. A datum is deemed as an outlier whenever it falls outside its 
associated confidence interval. To control the degree of smoothing effected by 
the 
estimator,~\cite{load_curve_cleansing} requires the user to label
 the outliers present in a training subset of data, and in this sense the 
approach therein is not 
fully automatic. Here instead, a novel alternative to load curve cleansing is 
developed after specializing the robust estimators
 of Sections \ref{sec:spacor} and \ref{sec:nonconvex}, to the case of cubic 
smoothing splines (Section \ref{ssec:real_data}). The smoothness-and outlier 
sparsity-controlling parameters are selected according to the guidelines
 in Section \ref{ssec:params}; hence, no input is required from the data 
analyst. The proposed spline-based method is tested on real load curve data 
from a government building.

Concluding remarks are given in Section 
\ref{sec:conclusion}, while some technical details are deferred to the 
Appendix.

\noindent\textit{Notation:} Bold uppercase letters will denote
matrices, whereas bold lowercase letters will stand for column
vectors. Operators $(\cdot)^{\prime}$, $\mbox{tr}(\cdot)$ and
$E[\cdot]$ will denote transposition, matrix trace and
expectation, respectively; $|\cdot|$ will be used for the
cardinality of a set and the magnitude of a scalar. The $\ell_q$ norm of vector
$\mathbf{x}\in\mathbb{R}^{p}$ is
$\|\mathbf{x}\|_q:=\left(\sum_{i=1}^p|x_i|^q\right)^{1/q}$ for
$q\geq 1$; and
$\|\mathbf{M}\|_F:=\sqrt{\mbox{tr}\left(\mathbf{M}\mathbf{M}^{\prime}\right)}$
is the matrix Frobenious norm. Positive definite matrices will
be denoted by $\bbM\succ\mathbf{0}$. The $p\times p$ identity
matrix will be represented by $\mathbf{I}_{p}$, while
$\mathbf{0}_{p}$ will denote the $p\times 1$ vector of all
zeros, and $\mathbf{0}_{p\times
q}:=\mathbf{0}_{p}\mathbf{0}_{q}^{\prime}$. 

\section{Robust Estimation Problem}\label{sec:model}

The training data comprises $N$ \textit{noisy} samples of $f$
taken at the input points $\{\mathbf{x}_i\}_{i=1}^{N}$ (also
known as knots in the splines parlance), and in the present
context they can be possibly contaminated with
outliers. Building on the parametric least-trimmed squares (LTS)
approach~\cite{rousseeuw}, the desired robust estimate
$\hat{f}$ can be obtained as the solution of the following
variational (V)LTS minimization problem
\begin{equation}\label{eq:variational_LTS}
\min_{f\in\mathcal{H}}\left[\sum_{i=1}^{s}r_{[i]}^2(f)+\mu\|f\|_{\mathcal{H}}^2\right]
\end{equation}
where $r_{[i]}^2(f)$ is the $i$-th order statistic among the
squared residuals $r_{1}^2(f),\ldots,r_{N}^2(f)$, and
$r_i(f):=y_i-f(\mathbf{x}_i)$. In words, given a feasible
$f\in\mathcal{H}$, to evaluate the sum of the cost in
\eqref{eq:variational_LTS} one: i) computes all $N$ squared
residuals $\{r_{i}^2(f)\}_{i=1}^{N}$, ii) orders them to form
the nondecreasing sequence $r_{[1]}^2(f)\leq\ldots\leq
r_{[N]}^2(f)$; and iii) sums up the smallest $s$ terms. As in the parametric 
LTS~\cite{rousseeuw}, the
so-termed trimming constant $s$ (also known as coverage) determines the 
breakdown point
of the VLTS estimator, since the largest $N-s$
residuals do not participate in \eqref{eq:variational_LTS}.
Ideally, one would like to make $N-s$ equal to the (typically
unknown) number of outliers $N_o$ in the training data. For most pragmatic 
scenaria where $N_o$ is unknown, the LTS estimator is an attractive option
due to its high breakdown point and desirable theoretical properties,
namely $\sqrt{N}$-consistency and asymptotic normality~\cite{rousseeuw}. 

The tuning
parameter $\mu\geq 0$ in \eqref{eq:variational_LTS} controls the tradeoff 
between fidelity to
the (trimmed) data, and the degree of ``smoothness'' measured
by $\|f\|_{\mathcal{H}}^2$. In particular, 
$\|f\|_{\mathcal{H}}^2$ can be interpreted as a generalized
ridge regularization term penalizing more those
functions with large coefficients in a basis expansion
involving the eigenfunctions of the kernel $K$.

Given that the sum in
\eqref{eq:variational_LTS} is a nonconvex functional, a
nontrivial issue pertains to the existence of the proposed VLTS
estimator, i.e., whether or not \eqref{eq:variational_LTS}
attains a minimum in $\mathcal{H}$. Fortunately, a
(conceptually) simple solution procedure suffices to show that
a minimizer does indeed exist.  Consider specifically a given
subsample of $s$ training data points, say
$\{y_i,\mathbf{x}_i\}_{i=1}^{s}$, and solve
\begin{equation*}
\min_{f\in\mathcal{H}}\left[\sum_{i=1}^{s}r_{i}^2(f)+\mu\|f\|_{\mathcal{H}}^2\right].
\end{equation*}
A unique minimizer of the form
$\hat{f}^{(j)}(\mathbf{x})=\sum_{i=1}^s\beta_{i}^{(j)}K(\mathbf{x},
\mathbf{x}_i)$
is guaranteed to exist, where $j$ is used here to denote the
chosen subsample, and the coefficients
$\{\beta_{i}^{(j)}\}_{i=1}^s$ can be obtained by solving a
particular linear system of equations~\cite[p. 11]{wahba}. This
procedure can be repeated for each subsample (there are
$J:={{N}\choose{s}}$ of these), to obtain a collection
$\{\hat{f}^{(j)}(\mathbf{x})\}_{j=1}^J$ of candidate solutions of
\eqref{eq:variational_LTS}. The winner(s) $\hat f:=\hat
f^{(j^\ast)}$ yielding the minimum cost, is the
desired VLTS estimator. A remark is now in order.

\begin{remark}[VLTS
complexity]\label{remark:VLTS_complexity} Even though conceptually
simple, the solution procedure just described guarantees
existence of (at least) one solution, but entails a
combinatorial search over all $J$ subsamples which is
intractable for moderate to large sample sizes $N$. In the context
of linear regression, algorithms to obtain
approximate LTS solutions are available; see e.g.,~\cite{LTS_algorithm}.
\end{remark}

\subsection{Robust function approximation via $\ell_0$-norm 
regularization}\label{ssec:LTS_equiv}

Instead of discarding large residuals, the alternative approach proposed here 
explicitly
 accounts for outliers in the regression model. To this end, consider the 
scalar
variables $\{o_i\}_{i=1}^{N}$ one per training datum,
taking the value $o_i=0$ whenever datum $i$ adheres to the postulated 
nominal model, and $o_i\neq 0$ otherwise. A regression model
naturally accounting for the presence of outliers is
\begin{equation}\label{eq:model_outliers}
y_{i}=f(\mathbf{x}_i)+o_i+\varepsilon_i, \quad\quad i=1,\ldots, N
\end{equation}
where $\{\varepsilon_i\}_{i=1}^N$ are zero-mean independent and
identically distributed (i.i.d.) random variables modeling the
observation errors. A similar model was advocated under
different assumptions in~\cite{fuchs} and~\cite{rao_robust}, in
the context of robust parametric regression; see 
also~\cite{candes_error_correction} and~\cite{Wright2010-IT}. For an 
outlier-free
datum $i$, \eqref{eq:model_outliers} reduces to
$y_{i}=f(\mathbf{x}_i)+\varepsilon_i$; hence, $\varepsilon_i$
will be often referred to as the nominal noise. Note that in
\eqref{eq:model_outliers}, both $f\in\mathcal{H}$ as well as
the $N\times 1$ vector $\mathbf{o}:=[o_1,\ldots,o_N]^\prime$
are unknown; thus, \eqref{eq:model_outliers} is
underdetermined. On the other hand, as outliers are expected to
often comprise a small fraction of the training sample say, not
exceeding 20\% -- vector $\mathbf{o}$ is typically
\textit{sparse}, i.e., most of its entries are zero; see also
Remark \ref{remark:sparsity_o}. Sparsity compensates for
underdeterminacy and provides valuable side-information when it
comes to efficiently estimating $\mathbf{o}$, identifying
outliers as a byproduct, and consequently performing
\textit{robust} estimation of the unknown function $f$. 

A natural criterion for
 controlling outlier sparsity is to seek the desired estimate $\hat{f}$ as 
the solution of
\begin{equation}\label{eq:cost_l0constr}
\min_{\substack{f\in\mathcal{H}\\\mathbf{o}\in\mathbb{R}^N}}
\left[\sum_{i=1}^{N}(y_{i}-f(\mathbf{x}_i)-o_i)^2+\mu\|f\|_{\mathcal{H}}^2\right],\quad
\textrm{ s.t. }\|\mathbf{o}\|_0\leq \tau
\end{equation}
where $\tau$ is a preselected threshold, and
$\|\mathbf{o}\|_0$ denotes the $\ell_0$-norm of
$\mathbf{o}$, which equals the number of nonzero entries of its
vector argument. 
Sparsity is directly controlled by the selection of the tuning parameter
$\tau\geq 0$. If the number of outliers $N_o$ were known a priori,
then $\tau$ should be selected equal to $N_o$. Unfortunately, 
analogously to related $\ell_0$-norm
constrained formulations in compressive sampling and sparse
signal representations, problem \eqref{eq:cost_l0constr} is
NP-hard. 
In addition, \eqref{eq:cost_l0constr} can be recast to an equivalent
(unconstrained) Lagrangian form; see
e.g.,~\cite{Bertsekas_Book_Nonlinear}
\begin{equation}\label{eq:cost_l0unconstr}
\min_{\substack{f\in\mathcal{H}\\\mathbf{o}\in\mathbb{R}^N}}
\left[\sum_{i=1}^{N}(y_{i}-f(\mathbf{x}_i)-o_i)^2
+\mu\|f\|_{\mathcal{H}}^2+\lambda_0\|\mathbf{o}\|_0\right]
\end{equation}
where the tuning Lagrange multiplier $\lambda_0\geq 0$ plays a role similar
to $\tau$ in \eqref{eq:cost_l0constr}, and the
$\ell_0$-norm sparsity encouraging penalty is added to
the cost. 

To further motivate model \eqref{eq:model_outliers}
and the proposed criterion \eqref{eq:cost_l0unconstr} for
robust nonparametric regression, it is worth checking the
structure of
the minimizers $\{\hat f,\hat{\mathbf{o}}\}$ of the cost in 
\eqref{eq:cost_l0unconstr}. Consider for the sake of argument 
 that $\lambda_0$ is given, 
and its value is such that  $\|\hat{\mathbf{o}}\|_0=\nu$, for
 some $0\leq\nu\leq N$. The goal is to characterize $\hat f$, as well as 
the
positions and values of the nonzero entries of $\hat{\mathbf{o}}$. Note 
that because $\|\hat{\mathbf{o}}\|_0=\nu$, the last term in 
\eqref{eq:cost_l0unconstr} is constant, hence inconsequential to the 
minimization. Upon defining $\hat{r}_i:=y_i-\hat{f}(\mathbf{x}_i)$, it is not 
hard to see that the entries of $\hat{\mathbf{o}}$ satisfy
\begin{equation}\label{eq:min_oi}
\hat{o}_i=\left\{\begin{array}{ccc}0,&&|\hat{r}_{i}|\leq \sqrt{\lambda_0}\\
\hat{r}_{i},&&|\hat{r}_{i}|>
\sqrt{\lambda_0}\end{array}\right.,\quad i=1,\ldots,N
\end{equation}
at the optimum. This is intuitive, since for those $\hat{o}_i\neq 0$ the best
 thing to do in 
terms of minimizing the overall cost is to set $\hat{o}_i=\hat{r}_{i}$, and 
thus null the corresponding squared-residual terms in 
\eqref{eq:cost_l0unconstr}. In conclusion, for the chosen value of $\lambda_0$ 
it holds that $\nu$ squared residuals effectively do not contribute to the 
cost in \eqref{eq:cost_l0unconstr}. 

To determine the support of 
$\hat{\mathbf{o}}$ and $\hat f$, one alternative is to exhaustively test all 
${{N}\choose{\nu}}$ admissible support combinations. For each one of these 
combinations (indexed by $j$), let 
$\mathcal{S}_j\subset\{1,\ldots,N\}$ be the index set describing the support of
 $\hat{\mathbf{o}}^{(j)}$, i.e., $\hat{o}_{i}^{(j)}\neq 0$ if and only if 
$i\in\mathcal{S}_j$; and $|\mathcal{S}_j|=\nu$. By virtue
 of \eqref{eq:min_oi}, the corresponding candidate $\hat{f}^{(j)}$ minimizes
\begin{equation*}
\min_{f\in\mathcal{H}}\left[\sum_{i\in\mathcal{S}_j}r_{i}^2(f)+\mu\|f\|_{
\mathcal{H}}^2\right]
\end{equation*}
while $\hat f$ is the one among all $\{\hat{f}^{(j)}\}$ that yields the least 
cost.
The previous discussion, in conjunction with the one preceding Remark 
\ref{remark:VLTS_complexity} completes
the argument required to establish the following result.

\begin{proposition}\label{prop:equiv}
If $\{\hat
f,\hat{\mathbf{o}}\}$ minimizes \eqref{eq:cost_l0unconstr} with
$\lambda_0$ chosen such that $\|\hat{\mathbf{o}}\|_0=N-s$, then
$\hat f$ also solves the VLTS problem \eqref{eq:variational_LTS}.
\end{proposition}

The importance of Proposition \ref{prop:equiv} is threefold. First, it
formally justifies model \eqref{eq:model_outliers} and its
estimator \eqref{eq:cost_l0unconstr} for robust function
approximation, in light of the well documented merits of LTS
regression~\cite{LTS_algorithm}. Second, it further solidifies the connection 
between
sparse linear regression and robust estimation. Third, the 
$\ell_0$-norm regularized formulation
in \eqref{eq:cost_l0unconstr} lends itself naturally to
efficient solvers based on convex relaxation, the
subject dealt with next.

\section{Sparsity Controlling Outlier Rejection}\label{sec:spacor}

To overcome the complexity hurdle in solving the
robust regression problem in
\eqref{eq:cost_l0unconstr}, one can resort to a suitable
relaxation of the objective function. The goal is to formulate
an optimization problem which is tractable, and whose solution
yields a satisfactory approximation to the minimizer of the
original hard problem. To this end, it is useful to recall that
the $\ell_1$-norm $\|\mathbf{x}\|_1$ of vector $\mathbf{x}$ is
the closest convex approximation of $\|\mathbf{x}\|_0$. This
property also utilized in the context of compressive
sampling~\cite{tropp_relax}, provides the motivation to relax
the NP-hard problem \eqref{eq:cost_l0unconstr} to
\begin{equation}\label{eq:cost_l1unconstr}
\min_{\substack{f\in\mathcal{H}\\\mathbf{o}\in\mathbb{R}^N}}
\left[\sum_{i=1}^{N}(y_{i}-f(\mathbf{x}_i)-o_i)^2
+\mu\|f\|_{\mathcal{H}}^2+\lambda_1\|\mathbf{o}\|_1\right].
\end{equation}
Being a convex optimization problem, \eqref{eq:cost_l1unconstr}
can be solved efficiently. The nondifferentiable $\ell_1$-norm
regularization term controls sparsity on the estimator of
$\mathbf{o}$, a property that has been recently exploited in
diverse problems in engineering, statistics and machine
learning. A noteworthy representative is the least-absolute
shrinkage and selection operator
(Lasso)~\cite{tibshirani_lasso}, a popular tool in statistics
for joint estimation and continuous variable selection in
linear regression problems. In its Lagrangian form, Lasso is
also known as basis pursuit denoising in the signal
processing literature, a term coined by~\cite{basis_pursuit} in
the context of finding the best sparse signal expansion using
an overcomplete basis.

It is pertinent to ponder on whether problem
\eqref{eq:cost_l1unconstr} has built-in ability to provide
robust estimates $\hat f$ in the presence of outliers. The
answer is in the affirmative, since a straightforward argument (details are 
deferred to 
the Appendix) shows that
\eqref{eq:cost_l1unconstr} is equivalent to a variational
M-type estimator found by
\begin{equation}\label{eq:variational_w_rho}
\min_{f\in\mathcal{H}}\left[\sum_{i=1}^{N}\rho(y_i-f(\mathbf{x}_i))+\mu\|f\|_{\mathcal{H}}^2\right]
\end{equation}
where $\rho:\mathbb{R}\rightarrow\mathbb{R}$ is a scaled
version of Huber's convex loss function~\cite{huber}
\begin{equation}\label{eq:rho_def}
\rho(u):=\left\{\begin{array}{ccc}u^2,&&|u|\leq\lambda_1/2\\
\lambda_1|u|-\lambda_1^2/4,&&|u|>\lambda_1/2\end{array}\right..
\end{equation}

\begin{remark}[Regularized regression and robustness]\label{remark:M-type_l1}
Existing works on linear regression have pointed out the
equivalence between $\ell_1$-norm
regularized regression and M-type estimators, under specific assumptions on the
 distribution of the outliers 
($\epsilon$-contamination)~\cite{fuchs,kekatos_gg_rs}. However,
they have not recognized the link with LTS through the convex
relaxation of \eqref{eq:cost_l0unconstr}, and the connection
asserted by Proposition \ref{prop:equiv}. Here, the treatment
goes beyond linear regression by considering nonparametric
functional approximation in RKHS. Linear regression is subsumed
as a special case, when the linear kernel
$K(\mathbf{x},\mathbf{y}):=\mathbf{x}^\prime\mathbf{y}$ is
adopted. In addition, no assumption is imposed on the outlier vector.
\end{remark}

It is interesting to compare the $\ell_0$- and $\ell_1$-norm
formulations  [cf. \eqref{eq:cost_l0unconstr} and
\eqref{eq:cost_l1unconstr}, respectively] in terms of their
equivalent purely variational counterparts in
\eqref{eq:variational_LTS} and
\eqref{eq:variational_w_rho}, that entail robust loss
functions. While the VLTS estimator completely discards large residuals,
$\rho$ still retains them, but downweighs their effect
through a linear penalty. Moreover, while \eqref{eq:variational_w_rho} is 
convex,
\eqref{eq:variational_LTS} is not and this has a direct impact on the 
complexity
to obtain either estimator. Regarding the trimming constant $s$ in 
\eqref{eq:variational_LTS}, it controls the number of residuals
retained and hence the breakdown point of VLTS. Considering instead
the threshold $\lambda_1/2$ in Huber's function $\rho$, when
the outliers' distribution is known a-priori, its value is
available in closed form so that the robust estimator is
optimal in a well-defined sense~\cite{huber}. Convergence in
probability of M-type cubic smoothing splines estimators -- a
special problem subsumed by \eqref{eq:variational_w_rho} -- was
studied in~\cite{cox}.

\subsection{Solving the convex relaxation}\label{ssec:solve_l1unconstr}

Because \eqref{eq:cost_l1unconstr} is jointly convex in $f$ and
$\mathbf{o}$, an alternating minimization (AM) algorithm  can
be adopted to solve \eqref{eq:cost_l1unconstr}, for fixed
values of $\mu$ and $\lambda_1$. Selection of these parameters
is a critical issue that will be discussed in Section
\ref{ssec:params}. AM solvers are iterative procedures that fix
one of the variables to its most up to date value, and minimize
the resulting cost with respect to the other one. Then the
roles are reversed to complete one cycle, and the overall
two-step minimization procedure is repeated for a prescribed
number of iterations, or, until a convergence criterion is met.
Letting $k=0,1,\ldots$ denote iterations, consider that
$\mathbf{o}:=\mathbf{o}^{(k-1)}$ is fixed in
\eqref{eq:cost_l1unconstr}. The update for $f^{(k)}$ at the
$k$-th iteration is given by
\begin{equation}\label{eq:update_f}
f^{(k)}:=\arg\min_{f\in\mathcal{H}}
\left[\sum_{i=1}^{N}\left((y_{i}-o_{i}^{(k-1)})-f(\mathbf{x}_i)\right)^2
+\mu\|f\|_{\mathcal{H}}^2\right]
\end{equation}
which corresponds to a standard regularization problem for
functional approximation in
$\mathcal{H}$~\cite{regnet_svm}, but with
\textit{outlier-compensated} data
$\left\{y_{i}-o_{i}^{(k-1)},\mathbf{x}_i\right\}_{i=1}^N$.
It is well known that the minimizer of the variational problem
\eqref{eq:update_f} is finitely parameterized, and given by the
kernel expansion
$f^{(k)}(\mathbf{x})=\sum_{i=1}^{N}\beta_i^{(k)}K(\mathbf{x},\mathbf{x}_i)$~\cite{wahba}.
The vector
$\bm\beta:=[\beta_1,\ldots,\beta_N]^\prime$ is found by solving
the linear system of equations
\begin{equation}\label{eq:betas}
\left[\mathbf{K}+\mu\mathbf{I}_N\right]\bm\beta^{(k)}=\mathbf{y}-\mathbf{o}^{(k-1)}
\end{equation}
where $\mathbf{y}:=[y_1,\ldots,y_N]^\prime$, and the $N\times
N$ matrix $\mathbf{K}\succ\mathbf{0}$ has entries
$[\mathbf{K}]_{ij}:=K(\mathbf{x}_i,\mathbf{x}_j)$.

In a nutshell, updating $f^{(k)}$ is equivalent to updating
vector $\bm\beta^{(k)}$ as per \eqref{eq:betas}, where only the
independent vector variable $\mathbf{y}-\mathbf{o}^{(k-1)}$
changes across iterations. Because the system matrix is
positive definite, the per iteration systems of linear
equations \eqref{eq:betas} can be efficiently solved after
computing once, the Cholesky factorization of
$\mathbf{K}+\mu\mathbf{I}_N$.

For fixed $f:=f^{(k)}$ in \eqref{eq:cost_l1unconstr}, the
outlier vector update $\mathbf{o}^{(k)}$ at iteration $k$ is
obtained as
\begin{equation}\label{eq:update_o}
\mathbf{o}^{(k)}:=\arg\min_{\mathbf{o}\in\mathbb{R}^N}
\left[\sum_{i=1}^{N}\left(r_{i}^{(k)}-o_{i}\right)^2
+\lambda_1\|\mathbf{o}\|_1\right]
\end{equation}
where
$r_{i}^{(k)}:=y_{i}-\sum_{j=1}^{N}\beta_j^{(k)}K(\mathbf{x}_i,\mathbf{x}_j)$.
Problem \eqref{eq:update_o} can be recognized as an instance of
Lasso for the so-termed orthonormal case, in particular
for an identity regression matrix. The solution of such Lasso
problems is readily obtained via
soft-thresholding~\cite{Friedman_Lasso_cd_2007}, in the form of
\begin{equation}\label{eq:update_ois}
o_i^{(k)}:=\mathcal{S}\left(r_{i}^{(k)},\lambda_1/2\right),\quad i=1,\ldots,N
\end{equation}
where $\mathcal{S}(z,\gamma):=\textrm{sign}(z)(|z|-\gamma)_+$
is the soft-thresholding operator, and
$(\cdot)_+:=\max(0,\cdot)$ denotes the projection onto the
nonnegative reals. The coordinatewise updates in
\eqref{eq:update_ois} are in par with the sparsifying property
of the $\ell_1$ norm, since for ``small'' residuals, i.e.,
$r_{i}^{(k)}\leq \lambda_1/2$, it follows that $o_i^{(k)}=0$, and the
$i$-th training datum is deemed outlier free. Updates
\eqref{eq:betas} and \eqref{eq:update_ois} comprise the
iterative AM solver of the $\ell_1$-norm regularized problem
\eqref{eq:cost_l1unconstr}, which is tabulated as Algorithm
\ref{table: AM_solver}. Convexity ensures convergence to the
global optimum solution regardless of the initial condition;
see e.g.,~\cite{Bertsekas_Book_Nonlinear}.

\begin{algorithm}[t]
\caption{: AM solver} \small{
\begin{algorithmic}
    \STATE Initialize $\mathbf{o}^{(-1)} =\mathbf{0}$, and run till
    convergence
    \FOR {$k=0,1$,$\ldots$}
        \STATE Update $\bm\beta^{(k)}$ solving $\left[\mathbf{K}+\mu\mathbf{I}_N\right]\bm\beta^{(k)}=
        \mathbf{y}-\mathbf{o}^{(k-1)}$.
        \STATE Update $\mathbf{o}^{(k)}$ via $o_i^{(k)}=\mathcal{S}\left(
        y_{i}-\sum_{j=1}^{N}\beta_j^{(k)}K(\mathbf{x}_i,\mathbf{x}_j),\lambda_1/2\right),\quad i=1,\ldots,N$.
    \ENDFOR
    \RETURN $f(\mathbf{x})=\sum_{i=1}^{N}\beta_i^{(\infty)}K(\mathbf{x},\mathbf{x}_i)$
\end{algorithmic}}
\label{table: AM_solver}
\end{algorithm}

Algorithm \ref{table: AM_solver} is also conceptually
interesting, since it explicitly reveals the intertwining
between the outlier identification process, and the estimation
of the regression function with the appropriate
outlier-compensated data. An additional point is worth
mentioning after inspection of \eqref{eq:update_ois} in the
limit as $k\to\infty$. From the definition of the
soft-thresholding operator $\mathcal S$, for those ``large''
residuals $\hat {r}_{i}:=\lim_{k\to\infty}r_{i}^{(k)}$
exceeding $\lambda_1/2$ in magnitude, $\hat{o}_i=\hat
{r}_{i}-\lambda_1/2$ when $\hat {r}_{i}>0$, and $\hat{o}_i=\hat
{r}_{i}+\lambda_1/2$ otherwise. In other words, larger
residuals that the method identifies as corresponding to
outlier-contaminated data are shrunk, but not completely
discarded. By plugging $\hat{\mathbf{o}}$ back into
\eqref{eq:cost_l1unconstr}, these ``large'' residuals cancel
out in the squared error term, but still contribute linearly
through the $\ell_1$-norm regularizer. This is exactly what one
would expect, in light of the equivalence established with the
variational $M$-type estimator in \eqref{eq:variational_w_rho}.

Next, it is established that an alternative to solving a
sequence of linear systems and scalar Lasso problems, is to
solve a single instance of the Lasso with specific response
vector and (non-orthonormal) regression matrix.

\begin{proposition}\label{prop:equiv_w_lasso}
Consider $\hat{\mathbf{o}}_{\textrm{Lasso}}$ defined as
\begin{equation}\label{eq:lasso}
\hat{\mathbf{o}}_{\textrm{Lasso}}:=\arg\min_{\mathbf{o}\in\mathbb{R}^{N}}\|
\mathbf{X}_\mu\mathbf{y}-\mathbf{X}_\mu\mathbf{o}\|_2^2+\lambda_1\|\mathbf{o}\|_1
\end{equation}
where
\begin{equation}\label{eq:X}
\mathbf{X}_\mu:=\left[\begin{array}{c}\mathbf{I}_N-\mathbf{K}\left(\mathbf{K}+\mu\mathbf{I}_N\right)^{-1}\\
\left(\mu\mathbf{K}\right)^{1/2}\left(\mathbf{K}+\mu\mathbf{I}_N\right)^{-1}\end{array}\right].
\end{equation}
Then the minimizers $\{\hat f,\hat{\mathbf{o}}\}$ of
\eqref{eq:cost_l1unconstr} are fully determined given
$\hat{\mathbf{o}}_{\textrm{Lasso}}$, as
$\hat{\mathbf{o}}:=\hat{\mathbf{o}}_{\textrm{Lasso}}$ and $\hat
f(\mathbf{x})=\sum_{i=1}^N\hat{\beta}_i
K(\mathbf{x},\mathbf{x}_i)$, with
$\hat{\bm\beta}=\left(\mathbf{K}+
\mu\mathbf{I}_N\right)^{-1}(\mathbf{y}-\hat{\mathbf{o}}_{\textrm{Lasso}})$.
\end{proposition}
\begin{IEEEproof}For notational convenience introduce the $N\times 1$ vectors
$\mathbf{f}:=[f(\mathbf{x}_1),\ldots,f(\mathbf{x}_N)]^\prime$
and $\hat{\mathbf{f}}:=[\hat f(\mathbf{x}_1),\ldots,\hat
f(\mathbf{x}_N)]^\prime$, where $\hat f\in\mathcal{H}$ is the
minimizer of \eqref{eq:cost_l1unconstr}. Next, consider
rewriting \eqref{eq:cost_l1unconstr} as
\begin{equation}\label{eq:iterated_l1unconstr}
\min_{\mathbf{o}\in\mathbb{R}^N}\left[\min_{f\in\mathcal{H}}
\left\|(\mathbf{y}-\mathbf{o})-\mathbf{f}\right\|_2^2
+\mu\|f\|_{\mathcal{H}}^2\right]+\lambda_1\|\mathbf{o}\|_1.
\end{equation}
The quantity inside the square brackets is a function of
$\mathbf{o}$, and can be written explicitly after carrying out
the minimization with respect to $f\in\mathcal{H}$. From the
results in~\cite{wahba}, it follows that the vector of optimum
predicted values at the points $\{\mathbf{x}_i\}_{i=1}^N$ is
given by
$\hat{\mathbf{f}}=\mathbf{K}\hat{\bm\beta}=\mathbf{K}\left(\mathbf{K}+
\mu\mathbf{I}_N\right)^{-1}(\mathbf{y}-\mathbf{o})$; see also
the discussion after \eqref{eq:update_f}. Similarly, one finds
that $\|\hat
f\|_{\mathcal{H}}^2=\hat{\bm\beta}^\prime\mathbf{K}\hat{\bm\beta}=
(\mathbf{y}-\mathbf{o})^\prime\left(\mathbf{K}+
\mu\mathbf{I}_N\right)^{-1}\mathbf{K}\left(\mathbf{K}+
\mu\mathbf{I}_N\right)^{-1}(\mathbf{y}-\mathbf{o})$. Having
minimized \eqref{eq:iterated_l1unconstr} with respect to $f$,
the quantity inside the square brackets is
$(\bm\Gamma_\mu:=\left(\mathbf{K}+
\mu\mathbf{I}_N\right)^{-1})$
\begin{align}\label{eq:simplification_steps}
\hspace{-0.3cm}\nonumber \min_{f\in\mathcal{H}}
\left[\left\|(\mathbf{y}-\mathbf{o})-\mathbf{f}\right\|_2^2
+\mu\|f\|_{\mathcal{H}}^2\right]& =\left\|(\mathbf{y}-\mathbf{o})-\hat{\mathbf{f}}\right\|_2^2
+\mu\|\hat f\|_{\mathcal{H}}^2\\
\nonumber& =\left\|(\mathbf{y}-\mathbf{o})-\mathbf{K}\bm\Gamma_\mu(\mathbf{y}-\mathbf{o})\right\|_2^2
+\mu(\mathbf{y}-\mathbf{o})^\prime\bm\Gamma_\mu\mathbf{K}\bm\Gamma_\mu(\mathbf{y}-\mathbf{o})\\
 & =\left\|(\mathbf{I}_N-\mathbf{K}\bm\Gamma_\mu)\mathbf{y}-(\mathbf{I}_N-\mathbf{K}\bm\Gamma_\mu)\mathbf{o}\right\|_2^2
+\mu(\mathbf{y}-\mathbf{o})^\prime\bm\Gamma_\mu\mathbf{K}\bm\Gamma_\mu(\mathbf{y}-\mathbf{o}).
\end{align}
After expanding the quadratic form in the right-hand side of
\eqref{eq:simplification_steps}, and eliminating the term that
does not depend on $\mathbf{o}$, problem
\eqref{eq:iterated_l1unconstr} becomes
\begin{equation*}
\min_{\mathbf{o}\in\mathbb{R}^N}\left[\left\|(\mathbf{I}_N-\mathbf{K}\bm\Gamma_\mu)\mathbf{y}-(\mathbf{I}_N-\mathbf{K}\bm\Gamma_\mu)\mathbf{o}\right\|_2^2
-2\mu\mathbf{y}^\prime\bm\Gamma_\mu\mathbf{K}\bm\Gamma_\mu\mathbf{o}
+\mu\mathbf{o}^\prime\bm\Gamma_\mu\mathbf{K}\bm\Gamma_\mu\mathbf{o}
+\lambda_1\|\mathbf{o}\|_1\right].
\end{equation*}
Completing the square one arrives at
\begin{equation*}
\min_{\mathbf{o}\in\mathbb{R}^N}\left[\left\|\left[\begin{array}{c}\mathbf{I}_N-\mathbf{K}\bm\Gamma_\mu\\
(\mu\mathbf{K})^{1/2}\bm\Gamma_\mu\end{array}\right]\mathbf{y}-
\left[\begin{array}{c}\mathbf{I}_N-\mathbf{K}\bm\Gamma_\mu\\
(\mu\mathbf{K})^{1/2}\bm\Gamma_\mu\end{array}\right]\mathbf{o}\right\|_2^2
+\lambda_1\|\mathbf{o}\|_1\right]
\end{equation*}
which completes the proof.
\end{IEEEproof}
The result in Proposition \ref{prop:equiv_w_lasso} opens the
possibility for effective methods to select $\lambda_1$. These
methods to be described in detail in the ensuing section,
capitalize on recent algorithmic advances on Lasso solvers,
which allow one to efficiently compute
$\hat{\mathbf{o}}_{\textrm{Lasso}}$ for all values of the
tuning parameter $\lambda_1$. This is crucial for obtaining
satisfactory robust estimates $\hat f$, since
\textit{controlling the sparsity} in $\mathbf{o}$ by tuning
$\lambda_1$ is tantamount to controlling the number of outliers
in model \eqref{eq:model_outliers}.

\subsection{Selection of the tuning parameters: robustification paths}\label{ssec:params}

As argued before, the tuning parameters $\mu$ and $\lambda_1$
in \eqref{eq:cost_l1unconstr} control the degree of smoothness
in $\hat f$ and the number of outliers (nonzero entries in
$\hat{\mathbf{o}}_{\textrm{Lasso}}$), respectively. From a
statistical learning theory standpoint, $\mu$ and $\lambda_1$
control the amount of regularization and model complexity, thus
capturing the so-termed effective degrees of
freedom~\cite{elements_of_statistics}. Complex models tend to
have worse generalization capability, even though the prediction
error over the training set $\mathcal{T}$ may be small
(overfitting). In the contexts of regularization
networks~\cite{regnet_svm} and Lasso estimation for
regression~\cite{tibshirani_lasso}, corresponding tuning
parameters are typically selected via model selection
techniques such as cross-validation, or, by minimizing the
prediction error over an independent test set, if
available~\cite{elements_of_statistics}. However, these simple
methods are severely challenged in the presence of multiple
outliers. For example, the \textit{swamping} effect refers to a
very large value of the residual $r_i$ corresponding to a left
out clean datum $\{y_i,\mathbf{x}_i\}$, because of an
unsatisfactory model estimation based on all data except $i$;
data which contain outliers.

The idea here offers an alternative method to overcome the
aforementioned challenges, and the possibility to efficiently
compute $\hat{\mathbf{o}}_{\textrm{Lasso}}$ for all values of
$\lambda_1$, given $\mu$. A brief overview of the state-of-the-art
in Lasso solvers is given first. Several methods for
selecting  $\mu$ and $\lambda_1$ are then described, which
differ on the assumptions of what is known regarding the
outlier model \eqref{eq:model_outliers}.

Lasso amounts to solving a quadratic programming (QP)
problem~\cite{tibshirani_lasso}; hence, an iterative procedure
is required to determine $\hat{\mathbf{o}}_{\textrm{Lasso}}$ in
\eqref{eq:lasso} for a given value of $\lambda_1$. While
standard QP solvers can be certainly invoked to this end, an
increasing amount of effort has been put recently toward
developing fast algorithms that capitalize on the unique
properties of Lasso. The LARS algorithm~\cite{Efron_lars_2004}
is an efficient scheme for computing the entire path of
solutions (corresponding to all values of $\lambda_1$),
sometimes referred to as regularization paths. LARS capitalizes
on piecewise linearity of the Lasso path of solutions, while incurring the 
complexity of a single
LS fit, i.e., when $\lambda_1=0$. Coordinate descent algorithms
have been shown competitive, even outperforming LARS when $p$
is large, as demonstrated in~\cite{friedman_2009}; see
also~\cite{Wu_Lasso_cd_2008,Friedman_Lasso_cd_2007}, and the
references therein. Coordinate descent solvers capitalize on
the fact that Lasso can afford a very simple solution in the
scalar case, which is given in closed form in terms
of a soft-thresholding operation [cf. \eqref{eq:update_ois}].
Further computational savings are attained through the use of
\textit{warm starts}~\cite{Friedman_Lasso_cd_2007}, when
computing the Lasso path of solutions over a grid of decreasing
values of $\lambda_1$. An efficient solver capitalizing on
variable separability has been proposed
in~\cite{Wright_tsp_2009}.

\begin{figure}[h]
\begin{center}
\includegraphics[width=\linewidth]{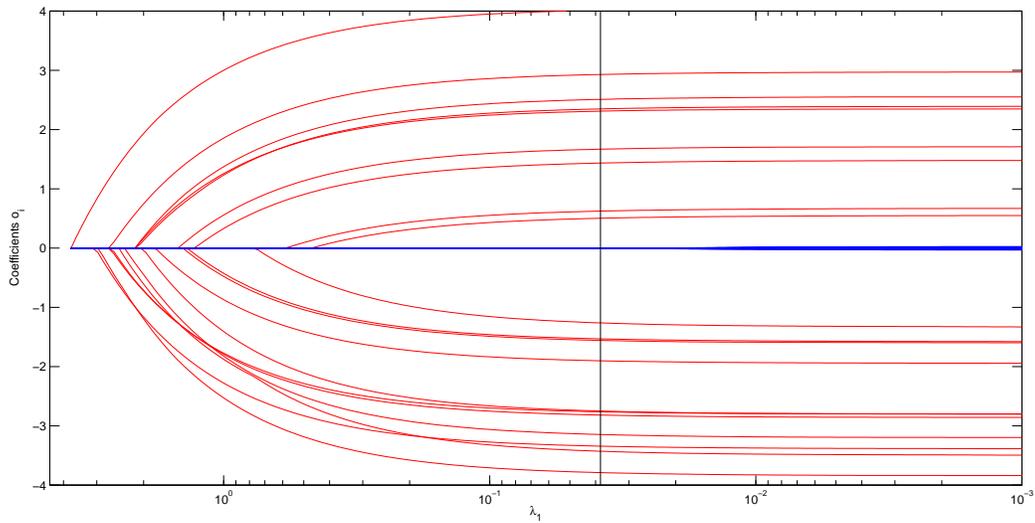}\\
\caption{Robustification path with optimum smoothing
parameter $\mu^{\ast}=1.55\times10^{-2}$. The data is corrupted with $N_o=20$
outliers. The coefficients $\hat{o}_i$ corresponding to the outliers are shown
in red, while the rest are shown in blue. The vertical line indicates the selection of
$\lambda_1^{\ast}=3.83\times10^{-2}$, and shows that the outliers were 
correctly
identified. }\label{fig:Fig_3}
\end{center}
\end{figure}

Consider then a grid of $G_\mu$ values of $\mu$ in the interval
$[\mu_{\min},\mu_{\max}]$, evenly spaced in a logarithmic
scale. Likewise, for each $\mu$ consider a similar type of grid
consisting of $G_\lambda$ values of $\lambda_1$, where
$\lambda_{\max}:=2\min_i|\mathbf{y}^\prime\mathbf{X}_\mu^\prime\mathbf{x}_{\mu,i}|$
is the minimum $\lambda_1$ value such that
$\hat{\mathbf{o}}_{\textrm{Lasso}}\neq\mathbf{0}_{N}$~\cite{friedman_2009},
and $\mathbf{X}_\mu:=[\mathbf{x}_{\mu,1}\ldots
\mathbf{x}_{\mu,N}]$ in \eqref{eq:lasso}. Typically,
$\lambda_{\min}=\epsilon\lambda_{\max}$ with
$\epsilon=10^{-4}$, say. Note that each of the $G_\mu$ values
of $\mu$ gives rise to a different $\lambda$ grid, since
$\lambda_{\max}$ depends on $\mu$ through $\mathbf{X}_\mu$.
Given the previously surveyed algorithmic alternatives to
tackle the Lasso, it is safe to assume that \eqref{eq:lasso}
can be efficiently solved over the (nonuniform) $G_\mu\times
G_\lambda$ grid of values of the tuning parameters. This way,
for each value of $\mu$ one obtains $G_\lambda$ samples of the
Lasso path of solutions, which in the present context can be
referred to as \textit{robustification path}. As $\lambda_1$
decreases, more variables $\hat{o}_{\textrm{Lasso},i}$ enter
the model signifying that more of the training data are deemed to contain 
outliers. An example of the robustification path
is given in Fig. \ref{fig:Fig_3}.

Based on the robustification paths and the prior knowledge
available on the outlier model \eqref{eq:model_outliers},
several alternatives are given next to select the ``best'' pair
$\{\mu,\lambda_1\}$ in the grid $G_\mu\times G_\lambda$.

\noindent\textit{Number of outliers is known:} When $N_o$ is
known, by direct inspection of the robustification paths one
can determine the range of values for $\lambda_1$, for which
$\hat{\mathbf{o}}_{\textrm{Lasso}}$ has exactly $N_o$ nonzero
entries. Specializing to the interval of interest, and after
discarding outliers which are now fixed and known, $K$-fold
cross-validation methods can be applied to determine
$\lambda_1$.

\noindent\textit{Variance of the nominal noise is known:} Supposing that the 
variance $\sigma_\varepsilon^2$ of the
i.i.d. nominal noise variables $\varepsilon_i$ in
\eqref{eq:model_outliers} is known, one can proceed as follows.
Using the solution $\hat f$ obtained for each pair
$\{\mu_i,\lambda_j\}$ on the grid, form the $G_\mu\times
G_\lambda$ sample variance matrix $\bar{\bm\Sigma}$ with
$ij$-th entry
\begin{equation}\label{eq:sample_variance}
[\bar{\bm\Sigma}]_{ij}:=\sum_{u|\hat{o}_{\textrm{Lasso},u}=0}\hat{r}_u^2/\hat
{N}_o=\sum_{u|\hat{o}_{\textrm{Lasso},u}=0}(y_u-\hat{f}(\mathbf{x}_u))^2/\hat
{N}_o
\end{equation}
where $\hat{N}_o$ stands for the number of nonzero entries in
$\hat{\mathbf{o}}_{\textrm{Lasso}}$. Although not made
explicit, the right-hand side of \eqref{eq:sample_variance}
depends on $\{\mu_i,\lambda_j\}$ through the estimate $\hat f$,
$\hat{\mathbf{o}}_{\textrm{Lasso}}$ and $\hat{N}_o$. The
entries $[\bar{\bm\Sigma}]_{ij}$ correspond to a sample
estimate of $\sigma_\varepsilon^2$, without considering those
training data $\{y_i,\mathbf{x}_i\}$ that the method
determined to be contaminated with outliers, i.e., those
indices $i$ for which $\hat{o}_{\textrm{Lasso},i}\neq0$. The
``winner'' tuning parameters $\{\mu^{\ast},\lambda_1^{\ast}\}:=
\{\mu_{i^{\ast}},\lambda_{j^{\ast}}\}$ are such that
\begin{equation}\label{eq:select_ij}
[i^{\ast},j^{\ast}]:=\arg\min_{i,j}|[\bar{\bm\Sigma}]_{ij}-\sigma_\varepsilon^2
|
\end{equation}
which is an absolute variance deviation (AVD) criterion.

\noindent\textit{Variance of the nominal noise is unknown:} If
$\sigma_\varepsilon^2$ is unknown, one can still compute a
robust estimate of the variance $\hat{\sigma}_\varepsilon^2$,
and repeat the previous procedure (with known nominal noise
variance) after replacing $\sigma_\varepsilon^2$ with
$\hat{\sigma}_\varepsilon^2$ in \eqref{eq:select_ij}. One
option is based on the median absolute deviation (MAD)
estimator, namely
\begin{equation}\label{eq:MAD}
\hat{\sigma}_\varepsilon:=1.4826\times\textrm{median}_i\left(|\hat 
r_i-\textrm{median}_j\left(\hat r_j\right)|\right)
\end{equation}
where the residuals $\hat r_i=y_i-\hat f(\mathbf{x}_i)$ are
formed based on a nonrobust estimate of $f$, obtained e.g.,
after solving \eqref{eq:cost_l1unconstr} with $\lambda_1=0$ and
using a small subset of the training dataset $\mathcal{T}$. The
factor $1.4826$ provides an approximately unbiased estimate of
the standard deviation when the nominal noise is Gaussian.
Typically, $\hat{\sigma}_\varepsilon$ in \eqref{eq:MAD} is used
as an estimate for the scale of the errors in general M-type
robust estimators; see e.g.,~\cite{cox} and~\cite{robust_svr}.

\begin{remark}[How sparse is sparse]\label{remark:sparsity_o}
Even though the very nature of outliers dictates that $N_o$ is
typically a small fraction of $N$ -- and thus $\mathbf{o}$ in
\eqref{eq:model_outliers} is sparse -- the method here
capitalizes on, but \textit{is not limited} to sparse settings.
For instance, choosing $\lambda_1\in[\lambda_{\min}\approx
0,\lambda_{\max}]$ along the robustification paths allows one
to continuously control the sparsity level, and potentially
select the right value of $\lambda_1$ for any given
$N_o\in\{1,\ldots,N\}$. Admittedly, if $N_o$ is large relative to $N$,
then even if it is possible to identify and discard the
outliers, the estimate $\hat f$ may not be accurate due to the
lack of outlier-free data.
\end{remark}

\section{Refinement via Nonconvex Regularization}\label{sec:nonconvex}

Instead of substituting $\|\mathbf{o}\|_0$ in
\eqref{eq:cost_l0unconstr} by its closest convex approximation,
namely $\|\mathbf{o}\|_1$, letting the surrogate function to be
non-convex can yield tighter approximations. For example, the
$\ell_0$-norm of a vector $\mathbf{x}\in\mathbb{R}^n$
was surrogated in~\cite{candes_l0_surrogate} by the logarithm
of the geometric mean of its elements, or by
$\sum_{i=1}^n\log|x_i|$. In rank minimization problems, apart
from the nuclear norm relaxation, minimizing the logarithm of
the determinant of the unknown matrix has been proposed as an
alternative surrogate~\cite{fazel_phdthesis}. Adopting
related ideas in the present nonparametric context, consider
approximating \eqref{eq:cost_l0unconstr} by
\begin{equation}\label{eq:cost_nonconvex}
\min_{\substack{f\in\mathcal{H}\\\mathbf{o}\in\mathbb{R}^N}}
\left[\sum_{i=1}^{N}(y_{i}-f(\mathbf{x}_i)-o_i)^2
+\mu\|f\|_{\mathcal{H}}^2+\lambda_0\sum_{i=1}^N\log(|o_i|+\delta)\right]
\end{equation}
where $\delta$ is a sufficiently small positive offset
introduced to avoid numerical instability.

Since the surrogate term in \eqref{eq:cost_nonconvex} is
concave, the overall problem is nonconvex. Still, local methods
based on iterative linearization of $\log(|o_i|+\delta)$,
around the current iterate $o_i^{(k)}$, can be adopted to
minimize \eqref{eq:cost_nonconvex}. From the concavity of the
logarithm, its local linear approximation serves as a global
overestimator. Standard majorization-minimization algorithms
motivate minimizing the global linear overestimator instead.
This leads to the following iteration for $k=0,1,\ldots$ (see
e.g.,~\cite{lange_surrogate} for further details)
\begin{align}
\label{eq:iters_nonconvex_fyo}[f^{(k)},\mathbf{o}^{(k)}]&:=\arg\min_{\substack{f\in\mathcal{H}\\\mathbf{o}\in\mathbb{R}^N}}
\left[\sum_{i=1}^{N}(y_{i}-f(\mathbf{x}_i)-o_i)^2
+\mu\|f\|_{\mathcal{H}}^2+\lambda_0\sum_{i=1}^N w_i^{(k)}|o_i|\right]\\
\label{eq:iters_nonconvex_w}w_i^{(k)}&:=\left(|o_i^{(k-1)}|+\delta\right)^{-1},\quad i=1,\ldots,N.
\end{align}
It is possible to eliminate the optimization variable
$f\in\mathcal{H}$ from \eqref{eq:iters_nonconvex_fyo}, by
direct application of the result in Proposition
\ref{prop:equiv_w_lasso}.  The equivalent update for
$\mathbf{o}$ at iteration $k$ is then given by
\begin{equation}\label{eq:iters_nonconvex_o}
\mathbf{o}^{(k)}:=\arg\min_{\mathbf{o}\in\mathbb{R}^N}
\left[\|
\mathbf{X}_\mu\mathbf{y}-\mathbf{X}_\mu\mathbf{o}\|_2^2+\lambda_0\sum_{i=1}^N w_i^{(k)}|o_i|\right]
\end{equation}
which amounts to an iteratively reweighted version of
\eqref{eq:lasso}. If the value of $|o_i^{(k-1)}|$ is small,
then in the next iteration the corresponding regularization
term $\lambda_0 w_i^{(k)}|o_i|$ has a large weight, thus
promoting shrinkage of that coordinate to zero. On the other
hand when $|o_i^{(k-1)}|$ is significant, the cost in the next
iteration downweighs the regularization, and places more
importance to the LS component of the fit. For small $\delta$,
analysis of the limiting point $\mathbf{o}^{\ast}$ of
\eqref{eq:iters_nonconvex_o} reveals that
\begin{equation*}
\lambda_0 w_i^{\ast}|o_i^{\ast}|\approx\left\{\begin{array}{cc}
\lambda_0, & |o_i^{\ast}|\neq 0\\
0, & |o_i^{\ast}|= 0
\end{array}\right.
\end{equation*}
and hence, $\lambda_0\sum_{i=1}^N
w_i^{\ast}|o_i^{\ast}|\approx\lambda_0\|\mathbf{o}^{\ast}\|_0$.

A good initialization for the iteration in
\eqref{eq:iters_nonconvex_o} and $\eqref{eq:iters_nonconvex_w}$
is  $\hat{\mathbf{o}}_{\textrm{Lasso}}$, which corresponds to
the solution of \eqref{eq:lasso} [and
\eqref{eq:cost_l1unconstr}] for $\lambda_0=\lambda_1^{\ast}$
and $\mu=\mu^{\ast}$. This is equivalent to a single iteration
of \eqref{eq:iters_nonconvex_o} with all weights equal to
unity. The numerical tests in Section \ref{sec:sims} will
indicate that even a single iteration of
\eqref{eq:iters_nonconvex_o} suffices to obtain improved
estimates $\hat f$, in comparison to those obtained from
\eqref{eq:lasso}. The following remark sheds further light
towards understanding why this should be expected.

\begin{remark}[Refinement through bias
reduction]\label{remark:bias_reduction} Uniformly weighted
$\ell_1$-norm regularized estimators such as
\eqref{eq:cost_l1unconstr} are biased~\cite{zou_adalasso}, due
to the shrinkage effected on the estimated coefficients. It
will be argued next that the improvements due to
\eqref{eq:iters_nonconvex_o} can be leveraged to bias
reduction. Several workarounds have been proposed to correct
the bias in sparse regression, that could as well be applied
here. A first possibility is to retain only the support of
\eqref{eq:lasso} and re-estimate the amplitudes via, e.g., the
unbiased LS estimator~\cite{Efron_lars_2004}. An alternative approach to 
reducing
bias is through nonconvex regularization using e.g., the
smoothly clipped absolute deviation (SCAD) scheme~\cite{scad}.
The SCAD penalty could replace the sum of logarithms in
\eqref{eq:cost_nonconvex}, still leading to a nonconvex
problem. To retain the efficiency of convex optimization
solvers while simultaneously limiting the bias, suitably
\textit{weighted} $\ell_1$-norm regularizers have been proposed
instead~\cite{zou_adalasso}. The constant weights
in~\cite{zou_adalasso} play a role similar to those in
\eqref{eq:iters_nonconvex_w}; hence, bias reduction is
expected.
\end{remark}

\section{Numerical Experiments}\label{sec:sims}

\subsection{Robust thin-plate smoothing
splines}\label{ssec:robust_splines}

To validate the proposed approach to robust nonparametric
regression, a simulated test is carried out here in the context
of thin-plate smoothing spline
approximation~\cite{duchon,wahba_weather}. Specializing
\eqref{eq:cost_l1unconstr} to this setup, the robust thin-plate
splines estimator can be formulated as
\begin{equation}\label{eq:thin_plate_splines}
\min_{\substack{f\in\mathcal{S}\\\mathbf{o}\in\mathbb{R}^N}}
\left[\sum_{i=1}^{N}(y_{i}-f(\mathbf{x}_i)-o_i)^2
+\mu\int_{\mathbb{R}^2}\|\nabla^2 f\|_F^2 d\mathbf{x}+\lambda_1\|\mathbf{o}\|_1\right]
\end{equation}
where  $||\nabla^2 f||_F$ denotes the Frobenius norm of the
Hessian of $f:\mathbb{R}^2\to\mathbb{R}$. The penalty
functional
\begin{equation}\label{eq:smoothing_penalty}
J[f]:=\int_{\mathbb{R}^2}\|\nabla^2 f\|_F^2 d\mathbf{x}=
\int_{\mathbb{R}^2}\left[\left(\frac{\partial^2 f}{\partial x_1^2}\right)^2
+2\left(\frac{\partial^2 f}{\partial x_1\partial x_2}\right)^2+
\left(\frac{\partial^2 f}{\partial x_2^2}\right)^2\right]d\mathbf{x}
\end{equation}
extends to $\mathbb{R}^2$ the one-dimensional roughness
regularization used in smoothing spline models. For $\mu=0$,
the (non-unique) estimate in \eqref{eq:thin_plate_splines}
corresponds to a \textit{rough} function interpolating the
outlier compensated data; while as $\mu\to\infty$ the estimate
is linear (cf. $\nabla^2 \hat
f(\mathbf{x})\equiv\mathbf{0}_{2\times 2}$). The optimization
is over $\mathcal{S}$, the space of Sobolev functions, for
which $J[f]$ is well defined \cite[p. 85]{duchon}. Reproducing
kernel Hilbert spaces such as $\mathcal{S}$, with
inner-products (and norms) involving derivatives are studied in
detail in~\cite{wahba}.

Different from the cases considered so far, the smoothing
penalty in \eqref{eq:smoothing_penalty} is only a seminorm,
since first-order polynomials vanish under $J[\cdot]$. Omitting
details than can be found in~\cite[p. 30]{wahba}, under fairly
general conditions a unique minimizer of
\eqref{eq:thin_plate_splines} exists. The solution admits the
finitely parametrized form
$\hat{f}(\mathbf{x})=\sum_{i=1}^N\beta_i
K(\mathbf{x},\mathbf{x}_i)+\bm\alpha_1^\prime
\mathbf{x}+\alpha_0$, where in this case
$K(\mathbf{x},\mathbf{y}):=\|\mathbf{x}-\mathbf{y}\|^2\log\|\mathbf{x}-\mathbf{y}\|$
is a radial basis function. In simple terms, the solution as a
kernel expansion is augmented with a member of the null space
of $J[\cdot]$. The unknown parameters
$\{\bm\beta,\bm\alpha_1,\alpha_0\}$ are obtained in closed
form, as solutions to a constrained, regularized LS problem;
see~\cite[p. 33]{wahba}. As a result, Proposition
\ref{prop:equiv_w_lasso} still holds with minor modifications
on the structure of $\mathbf{X}_\mu$.
\begin{remark}[Bayesian framework]\label{remark:bayes}
Adopting a Bayesian perspective, one could model $f(\mathbf{x})$ in 
\eqref{eq:model_outliers} as a sample function of a zero mean Gaussian 
stationary process, with covariance function 
$K(\mathbf{x},\mathbf{y})=\|\mathbf{x}-\mathbf{y}\|^2\log\|\mathbf{x}-\mathbf{
y}\|$~\cite{splines_bayes}. Consider as well that 
$\{f(\mathbf{x}),\{o_i,\varepsilon_i\}_{i=1}^N\}$ are mutually independent, 
while $\varepsilon_i\sim\mathcal{N}(0,\mu^\ast/2)$ and 
$o_i\sim\mathcal{L}(0,\mu^\ast/\lambda_1^\ast)$ in \eqref{eq:model_outliers}  
are i.i.d. Gaussian and Laplace
 distributed, respectively. From the results in~\cite{splines_bayes} and a 
straightforward calculation, it follows that setting $\lambda_1=\lambda_1^\ast$
 and $\mu=\mu^\ast$ in \eqref{eq:thin_plate_splines} yields estimates $\hat f$ 
(and $\hat{\mathbf{o}}$) which are optimal in a maximum a posteriori sense. 
This provides yet another means of selecting the parameters $\mu$ and 
$\lambda_1$, further expanding the options presented in Section 
\ref{ssec:params}.
\end{remark}

\begin{figure}[h] 
\vspace{-0.5cm}
\begin{center}
\includegraphics[width=0.9\linewidth]{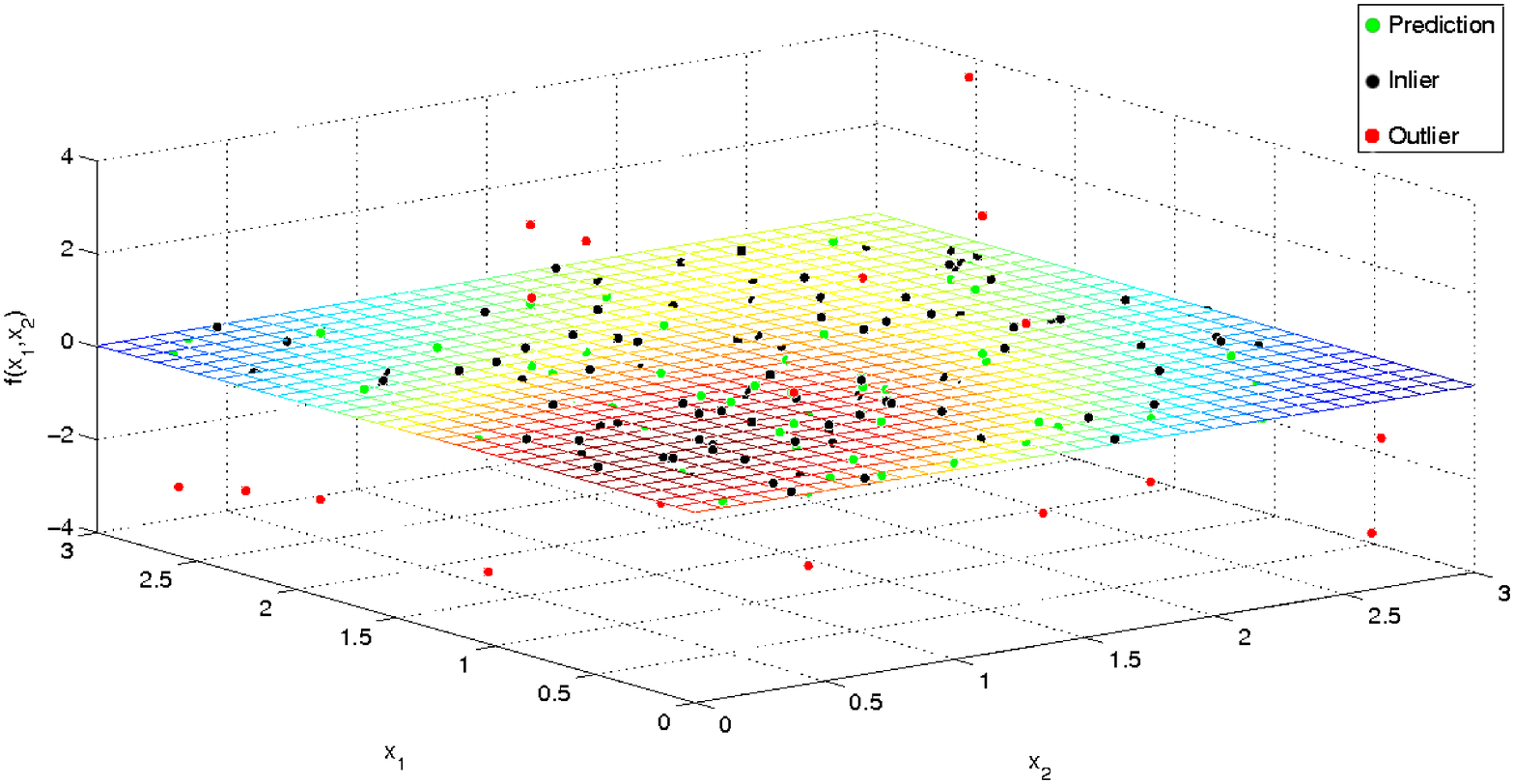}\\
\vspace{-0.5cm} 
\caption{True Gaussian mixture function $f_o(\mathbf{x})$,
and its $180$ noisy samples taken over $[0,3]\times[0,3]$ shown as black dots.
The red dots indicate the $N_o=20$ outliers in the training data set $\mathcal{T}$.
The green points indicate the predicted responses $\hat y_i$ at the sampling points $\mathbf{x}_i$,
from the estimate $\hat f$ obtained after solving \eqref{eq:thin_plate_splines}.
Note how all green points are close to the surface $f_o$.}\label{fig:Fig_2}
\end{center}
\end{figure}

\begin{figure}[h]
\begin{minipage}[b]{0.5\linewidth}
  \centering
  \centerline{\includegraphics[width=\linewidth]{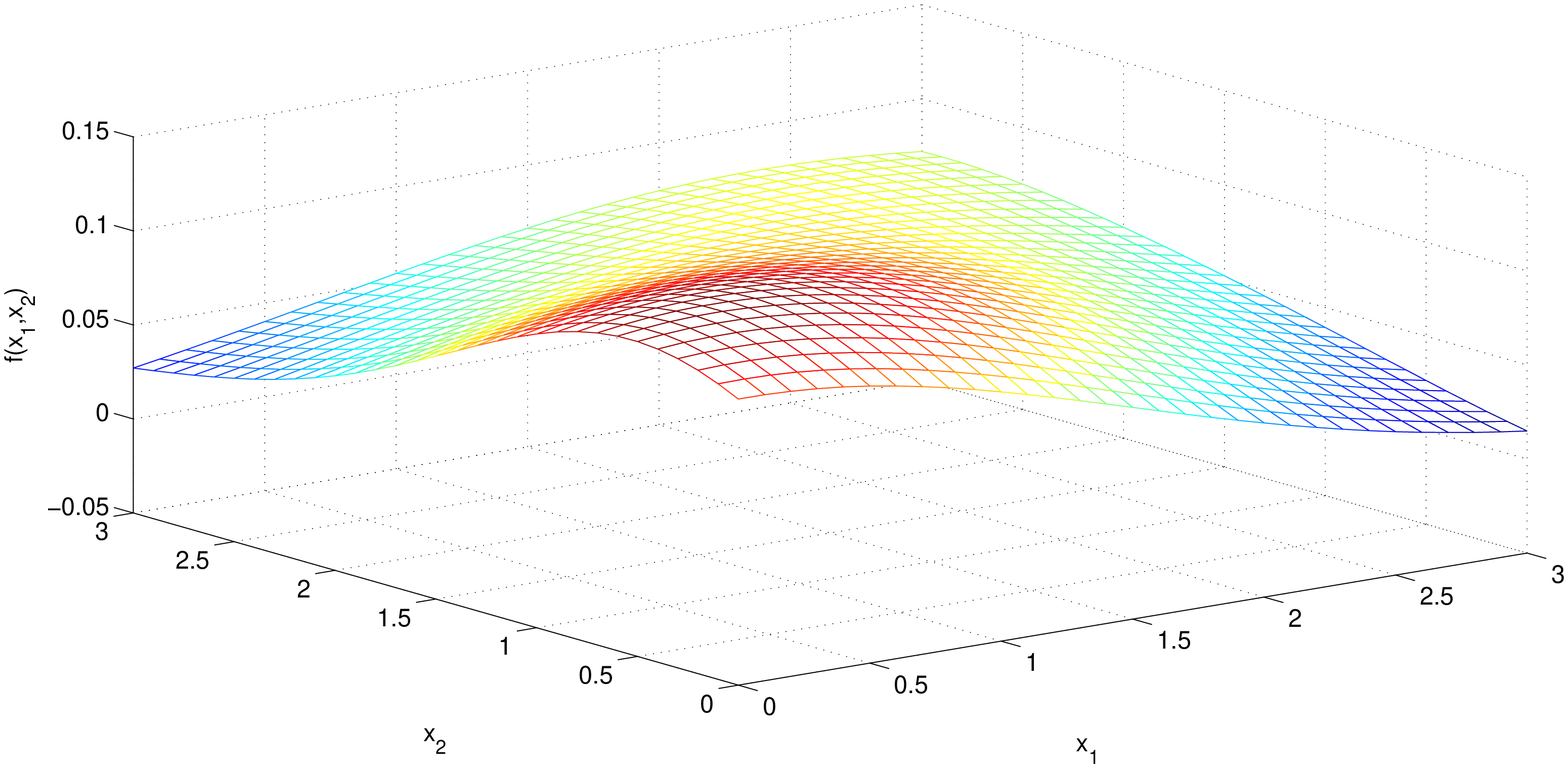}}
\centerline{(a)} 
\end{minipage}
\hfill
\begin{minipage}[b]{.5\linewidth}
  \centering
  \centerline{\includegraphics[width=\linewidth]
  {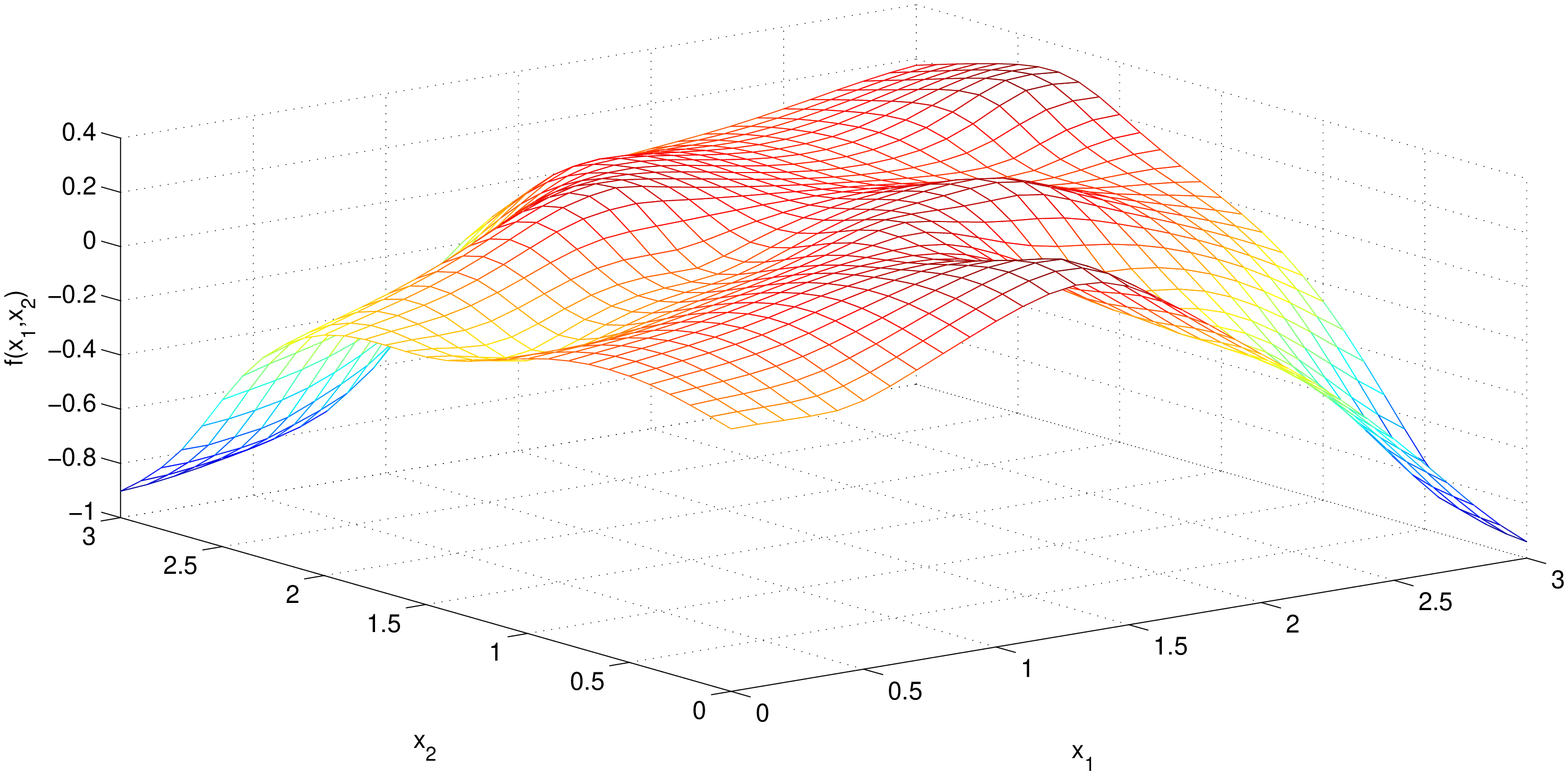}}
\centerline{(b)} 
\end{minipage}\\
\begin{minipage}[b]{0.5\linewidth}
  \centering
  \centerline{\includegraphics[width=\linewidth]{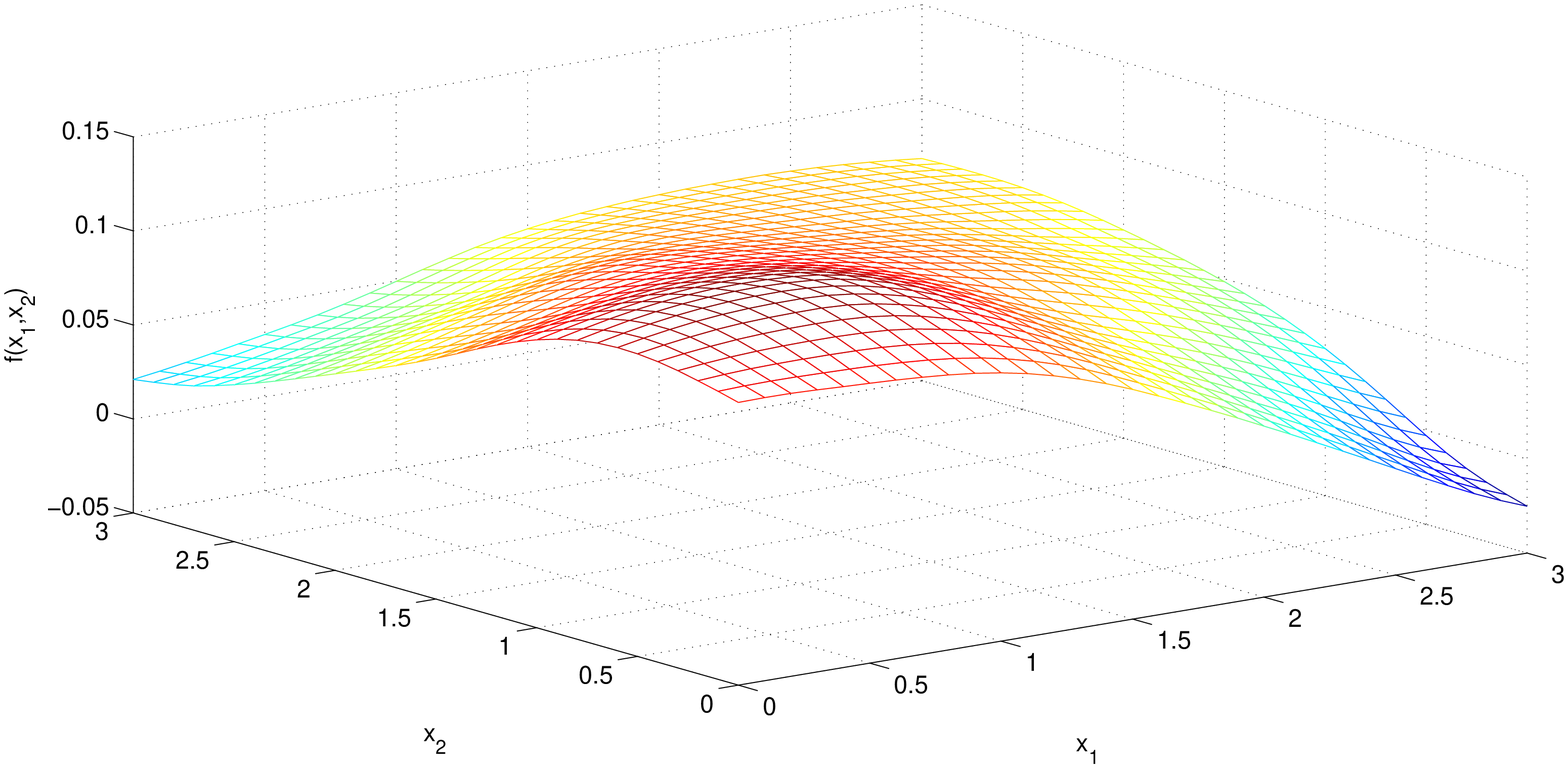}}
\centerline{(c)} 
\end{minipage}
\hfill
\begin{minipage}[b]{.5\linewidth}
  \centering
  \centerline{\includegraphics[width=\linewidth]
  {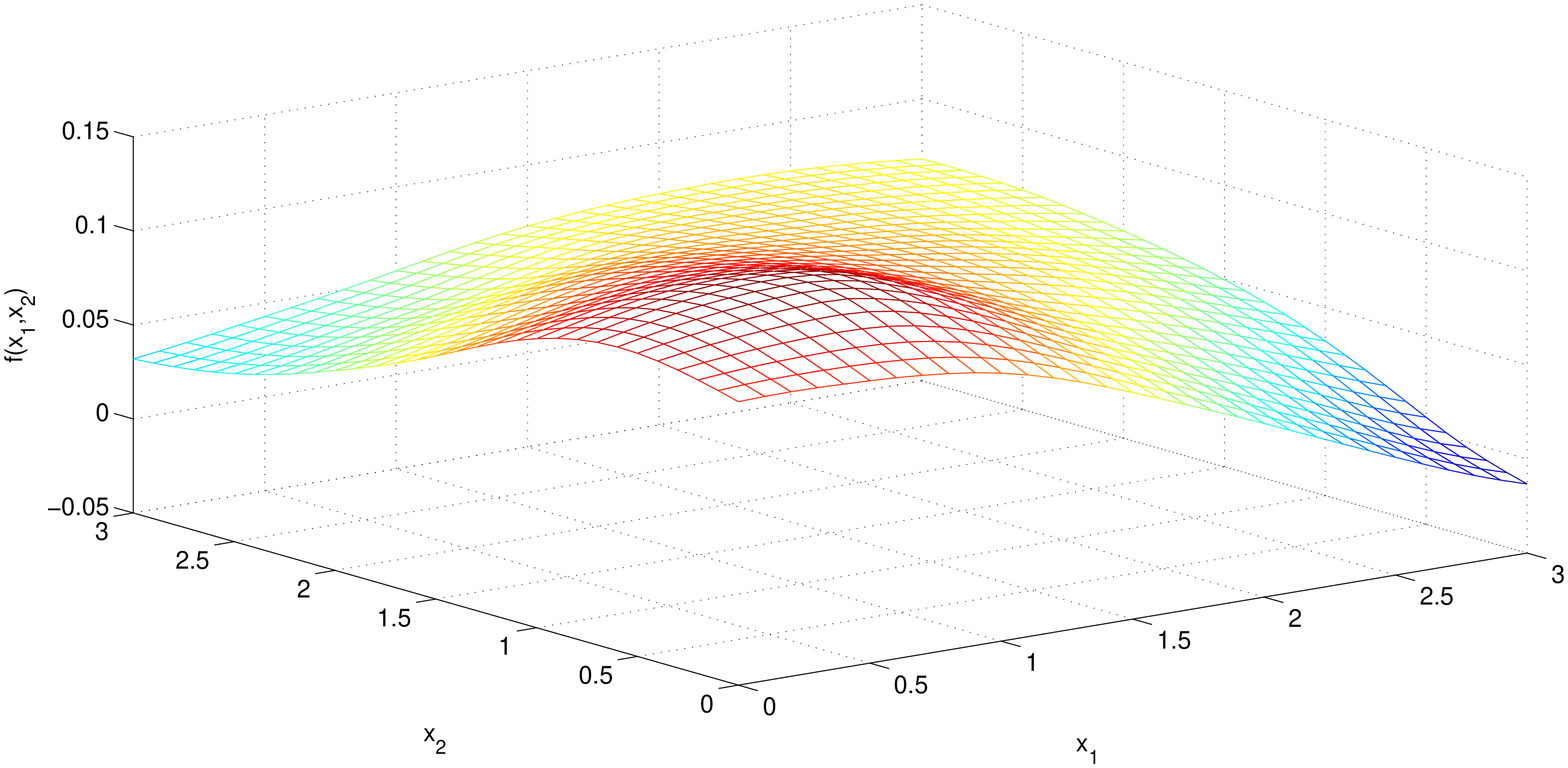}}
\centerline{(d)} 
\end{minipage}
\caption{Robust estimation of a Gaussian mixture using
thin-plate splines. The data is corrupted with $N_o=20$
outliers. (a) True function $f_o(\mathbf{x})$;
(b) nonrobust predicted function obtained after solving
\eqref{eq:nr_thin_plate_splines};
(c) predicted function after solving
\eqref{eq:thin_plate_splines} with the optimum tuning
parameters; (d) refined predicted function using the
nonconvex regularization in \eqref{eq:cost_nonconvex}.}
\label{fig:Fig_1}
\end{figure}

The simulation setup is as follows. Noisy samples of the true
function $f_o:\mathbb{R}^2\to\mathbb{R}$ comprise the training
set $\mathcal{T}$. Function $f_o$ is generated as a Gaussian
mixture with two components, with respective mean vectors and
covariance matrices given by
\begin{equation*}
\bm\mu_1=\left[\begin{array}{c}0.2295\\0.4996\end{array}\right],
\:\bm\Sigma_1=\left[\begin{array}{cc}2.2431&0.4577\\
0.4577&1.0037\end{array}\right],\quad
\bm\mu_2=\left[\begin{array}{c}2.4566\\2.9461\end{array}\right],
\:\bm\Sigma_2=\left[\begin{array}{cc}2.9069&0.5236\\
0.5236&1.7299\end{array}\right].
\end{equation*}
Function $f_o(\mathbf{x})$ is depicted in Fig. \ref{fig:Fig_1}
(a). The training data set comprises $N=200$ examples,
with inputs $\{\mathbf{x}_i\}_{i=1}^N$ drawn from a uniform
distribution in the square $[0,3]\times[0,3]$. Several values
ranging from $5\%$ to $25\%$ of the data are generated
contaminated with outliers. Without loss of generality, the
corrupted data correspond to the first $N_o$ training samples
with $N_o=\{10,20,30,40,50\}$, for which the response values
$\{y_i\}_{i=1}^{N_o}$ are independently drawn from a uniform
distribution over $[-4,4]$. Outlier-free data are generated according to the
model $y_i=f_o(\mathbf{x}_i)+\varepsilon_i$, where the
independent additive noise terms
$\varepsilon_i\sim\mathcal{N}(0,10^{-3})$ are Gaussian
distributed, for $i=N_o+1,\ldots,200$. For the case where
$N_o=20$, the data used in the experiment is shown in Fig.
\ref{fig:Fig_2}. Superimposed to the true function $f_o$ are
$180$ black points corresponding to data drawn from the nominal model, as
 well as
$20$ red outlier points.

For this experiment, the nominal noise variance
$\sigma_\varepsilon^2=10^{-3}$ is assumed known. A nonuniform
grid of $\mu$ and $\lambda_1$ values is constructed, as
described in Section \ref{ssec:params}. The relevant parameters
are $G_{\mu}=G_{\lambda}=200$, $\mu_{\min}=10^{-9}$ and
$\mu_{\max}=1$. For each value of $\mu$, the $\lambda_1$ grid
spans the interval defined by
$\lambda_{\max}:=2\min_i|\mathbf{y}^\prime\mathbf{X}_\mu^\prime\mathbf{x}_{\mu,i}|$
and $\lambda_{\min}=\epsilon\lambda_{\max}$, where
$\epsilon=10^{-4}$. Each of the $G_{\mu}$ robustification paths
corresponding to the solution  of \eqref{eq:lasso} is obtained
using the SpaRSA toolbox in~\cite{Wright_tsp_2009},
exploiting warm starts for faster convergence.  Fig.
\ref{fig:Fig_3} depicts an example with $N_o=20$ and
$\mu^{\ast}=1.55\times 10^{-2}$. With the robustification paths
at hand, it is possible to form the sample variance matrix
$\bar{\bm\Sigma}$ [cf. \eqref{eq:sample_variance}], and select
the optimum tuning parameters $\{\mu^{\ast},\lambda_1^{\ast}\}$
based on the criterion \eqref{eq:select_ij}. Finally, the
robust estimates are refined by running a single iteration of
\eqref{eq:iters_nonconvex_o} as described in Section
\ref{sec:nonconvex}. The value $\delta=10^{-5}$ was utilized,
and several experiments indicated that the results are quite
insensitive to the selection of this parameter.

\begin{table}[t]
\renewcommand{\arraystretch}{1.3}
\caption{Results for the thin-plate splines simulated test}
\label{table:results} \centering
\begin{tabular}{|c|c|c|c|c|c|c|}
\hline
\bfseries $N_o$ & \bfseries $\lambda_1^{\ast}$ &  \bfseries $\mu^{\ast}$ & 
\bfseries $\bar{\textrm{err}}$ for
\eqref{eq:cost_l1unconstr}& \bfseries $\bar{\textrm{err}}$ for
\eqref{eq:cost_nonconvex}& \bfseries $\textrm{Err}_{\mathcal{T}}$ for
\eqref{eq:cost_l1unconstr}& \bfseries $\textrm{Err}_{\mathcal{T}}$ for
\eqref{eq:cost_nonconvex}\\
\hline\hline
$10$ & $3.87\times10^{-2}$ & $2.90\times10^{-3}$ & $1.00\times 10^{-4}$ & 
$1.03\times 10^{-4}$ & $2.37\times 10^{-5}$ & $2.27\times 10^{-5}$\\\hline
$20$ & $3.83\times10^{-2}$ & $1.55\times10^{-2}$ & $1.00\times 10^{-4}$ & 
$9.16\times 10^{-5}$ & $4.27\times 10^{-5}$ & $2.39\times 10^{-5}$\\\hline
$30$ & $2.28\times10^{-2}$ & $6.67\times10^{-2}$ & $1.22\times 10^{-4}$ & 
$1.18\times 10^{-4}$ & $2.89\times 10^{-5}$ & $1.93\times 10^{-5}$\\\hline
$40$ & $2.79\times10^{-2}$ & $6.10\times10^{-3}$ & $1.01\times 10^{-4}$ & 
$1.14\times 10^{-4}$ & $1.57\times 10^{-5}$ & $1.32\times 10^{-5}$\\\hline
$50$ & $2.49\times10^{-2}$ & $5.42\times10^{-2}$ & $1.01\times 10^{-4}$ & 
$9.9\times 10^{-5}$ & $1.19\times 10^{-5}$ & $1.05\times 10^{-5}$\\
\hline
\end{tabular}
\end{table}

The same experiment was conducted for a variable number of
outliers $N_o$, and the results are listed in Table
\ref{table:results}. In all cases, a $100\%$ outlier
identification success rate was obtained, for the chosen value
of the tuning parameters. This even happened at the first stage
of the method, i.e., $\hat{\mathbf{o}}_{\textrm{Lasso}}$ in
\eqref{eq:lasso} had the correct support in all cases. It has
been observed in some other setups that \eqref{eq:lasso} may
select a larger support than $[1,N_o]$, but after running a few
iterations of \eqref{eq:iters_nonconvex_o} the true support was
typically identified. To assess quality of the estimated
function $\hat f$, two figures of merit were considered. First,
the \textit{training error} $\bar{\textrm{err}}$ was evaluated
as
\begin{equation*}
\bar{\textrm{err}}=\frac{1}{N-N_o}\sum_{i=N_o}^{N}\left(y_i-\hat{f}(\mathbf{x}_i)\right)^2
\end{equation*}
i.e., the average loss over the training sample $\mathcal{T}$
after excluding outliers. Second, to assess the generalization
capability of $\hat f$, an approximation to the
\textit{generalization error} $\textrm{Err}_{\mathcal{T}}$ was
computed as
\begin{equation}\label{eq:gen_error}
\textrm{Err}_{\mathcal{T}}=E\left[\left(y-\hat{f}(\mathbf{x})\right)^2|\mathcal{T}\right]
\approx\frac{1}{\tilde{N}}\sum_{i=1}^{\tilde{N}}\left(\tilde{y}_i-\hat{f}(\tilde{\mathbf{x}}_i)\right)^2
\end{equation}
where $\{\tilde{y}_i,\tilde{\mathbf{x}}_i\}_{i=1}^{\tilde{N}}$
is an independent test set generated from the model
$\tilde{y}_i=f_o(\tilde{\mathbf{x}}_i)+\varepsilon_i$. For the
results in Table \ref{table:results}, $\tilde{N}=961$ was
adopted corresponding to a uniform rectangular grid of $31
\times 31$ points $\tilde{\mathbf{x}}_i$ in $[0,3]\times
[0,3]$. Inspection of Table \ref{table:results} reveals that
the training errors $\bar{\textrm{err}}$ are comparable for the
function estimates obtained after solving
\eqref{eq:cost_l1unconstr} or its nonconvex refinement
\eqref{eq:cost_nonconvex}. Interestingly, when it comes to the
more pragmatic generalization error
$\textrm{Err}_{\mathcal{T}}$, the refined estimator
\eqref{eq:cost_nonconvex} has an edge for all values of $N_o$.
As expected, the bias reduction effected by the iteratively
reweighting procedure of Section \ref{sec:nonconvex} improves
considerably the generalization capability of the method; see
also Remark \ref{remark:bias_reduction}.

A pictorial summary of the results is given in Fig
\ref{fig:Fig_1}, for $N_o=20$ outliers. Fig \ref{fig:Fig_1}
(a) depicts the true Gaussian mixture $f_o(\mathbf{x})$,
whereas Fig. \ref{fig:Fig_1} (b) shows the nonrobust
thin-plate splines estimate obtained after solving
\begin{equation}\label{eq:nr_thin_plate_splines}
\min_{f\in\mathcal{S}}
\left[\sum_{i=1}^{N}(y_{i}-f(\mathbf{x}_i))^2
+\mu\int_{\mathbb{R}^2}\|\nabla^2 f\|_F^2 d\mathbf{x}\right].
\end{equation}
Even though the thin-plate penalty enforces some degree of smoothness, the
estimate is severely disrupted by the presence of outliers [cf.
the difference on the $z$-axis ranges]. On the other hand, Figs.
\ref{fig:Fig_1} (c) and (d), respectively, show the
robust estimate $\hat f$ with
$\lambda_1^{\ast}=3.83\times10^{-2}$, and its bias reducing
refinement. The improvement is apparent, corroborating the
effectiveness of the proposed approach.

\subsection{Sinc function estimation}\label{ssec:sinc_approx}

\begin{table}[t]
\renewcommand{\arraystretch}{1.3}
\caption{Generalization error ($\textrm{Err}_{\mathcal{T}}$) results for the 
$\textrm{sinc}$ function estimation experiment}
\label{table:results_sinc} \centering
\begin{tabular}{|c|c|c|c|}
\hline
\bfseries Method &  $\sigma_\varepsilon^2=1\times 10^{-4}$ &
$\sigma_\varepsilon^2=1\times 10^{-3}$ &  $\sigma_\varepsilon^2=1\times 
10^{-2}$ \\
\hline\hline
\bfseries Nonrobust [\eqref{eq:regnets} 
with $V(u)=u^2$]  & $5.67\times10^{-2}$ & $8.28\times10^{-2}$ & $1.13\times 
10^{-1}$ \\\hline
\bfseries SVR with $\epsilon=0.1$ & $5.00\times10^{-3}$ & $6.42\times10^{-4}$ &
 $6.15\times 
10^{-3}$ \\\hline
\bfseries 
RSVR  with
$\epsilon=0.1$ & $1.10\times10^{-3}$ & $5.10\times10^{-4}$ & $4.47\times 
10^{-3}$ \\\hline
\bfseries 
\bfseries SVR with  
$\epsilon=0.01$ & $8.24\times10^{-5}$ & $4.79\times10^{-4}$ & $5.60\times 
10^{-3}$ \\\hline
\bfseries RSVR with
$\epsilon=0.01$ & $7.75\times10^{-5}$ & $3.90\times10^{-4}$ & $3.32\times 
10^{-3}$ \\\hline
\bfseries Sparsity-controlling in \eqref{eq:cost_l1unconstr} & 
$1.47\times10^{-4}$ & 
$6.56\times10^{-4}$ & $4.60\times 
10^{-3}$ \\\hline
\bfseries Refinement in \eqref{eq:cost_nonconvex} & $7.46\times10^{-5}$ & 
$3.59\times10^{-4}$
 & $3.21\times 
10^{-3}$ \\\hline
\end{tabular}
\end{table}

The univariate function $\textrm{sinc}(x):=\textrm{sin}(\pi x)/(\pi x)$ is 
commonly 
adopted to evaluate the performance of nonparametric regression 
methods~\cite{RSVR,robust_kernel_regression}. Given noisy training examples 
with a small fraction of outliers,
approximating $\textrm{sinc}(x)$ over the interval $[-5,5]$ is considered in
 the present simulated test. The 
sparsity-controlling robust nonparametric regression methods of this paper 
are compared with the SVR~\cite{Vapnik_Learning_Theory} and robust 
SVR in~\cite{RSVR}, for the case of the $\epsilon$-insensitve loss 
function with values $\epsilon=0.1$ and $\epsilon=0.01$. In 
order to implement (R)SVR, routines from a publicly available SVM Matlab 
toolbox 
were utilized~\cite{SVR_Matlab}. Results for the nonrobust regularization 
network approach in \eqref{eq:regnets} (with $V(u)=u^2$) are reported as well, 
to assess the performance degradation incurred when compared to the aforementioned robust alternatives. 
Because the fraction of outliers ($N_o/N$) in the training data is
assumed known to the method of~\cite{RSVR}, the same will be assumed
towards selecting the tuning parameters $\lambda_1$ and $\mu$ in \eqref{eq:cost_l1unconstr}, 
as described in Section \ref{ssec:params}. The 
$\{\mu,\lambda_1\}$-grid parameters 
selected for the experiment in Section \ref{ssec:robust_splines} were used here
 as well, except for $\mu_{\min}=10^{-5}$.  Space $\mathcal{H}$ is chosen
to be the RKHS induced by the positive definite Gaussian kernel function 
$K(u,v)=\textrm{exp}\left[-(u-v)^2/(2\eta^2)\right]$,
with parameter $\eta=0.1$ for all cases.

\begin{figure}[h]
\begin{minipage}[b]{0.4\linewidth}
  \centering
  \centerline{\includegraphics[width=\linewidth]{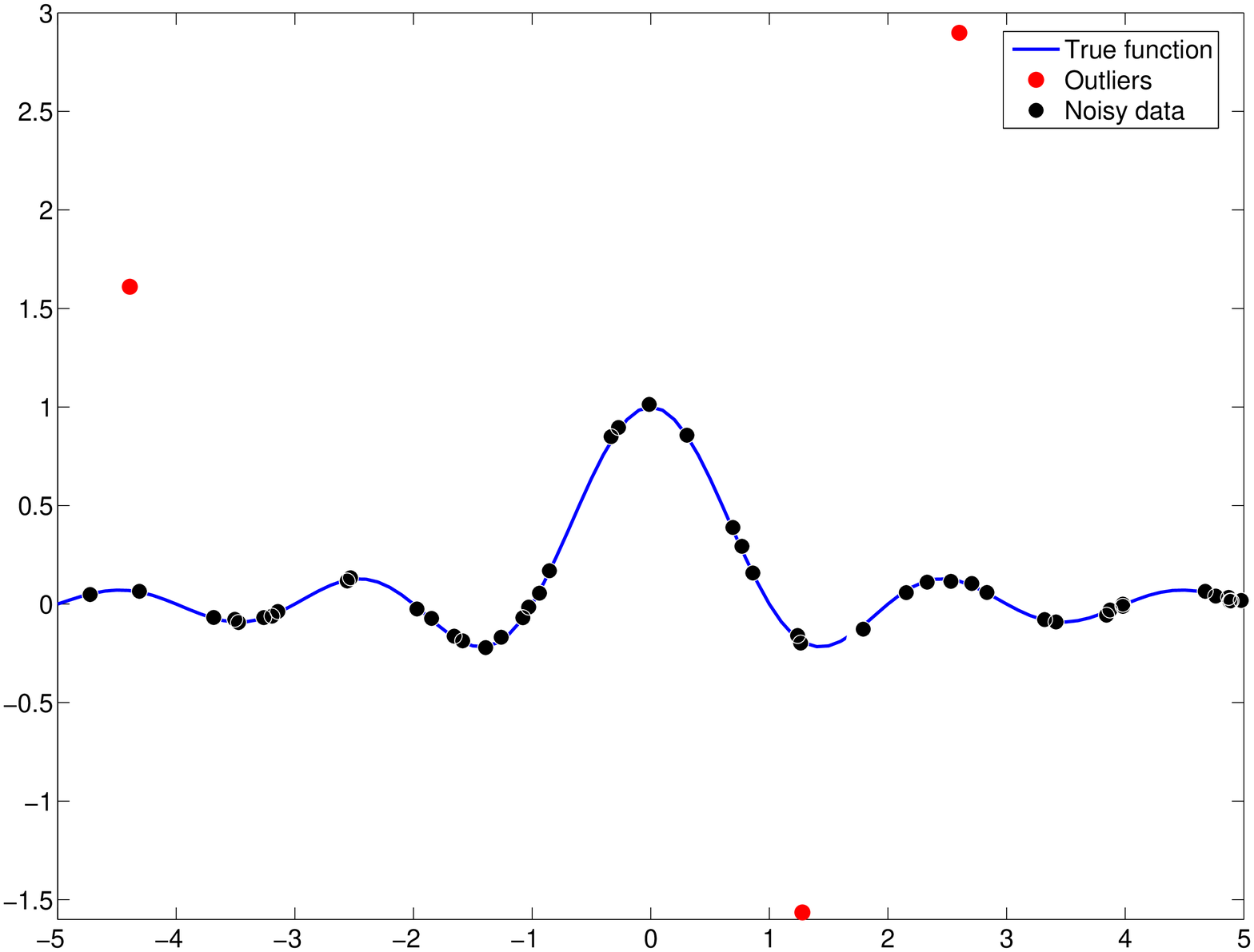}}
\vspace{-0.5cm}\centerline{(a)} 
\end{minipage}
\hfill
\begin{minipage}[b]{.4\linewidth}
  \centering
  \centerline{\includegraphics[width=\linewidth]{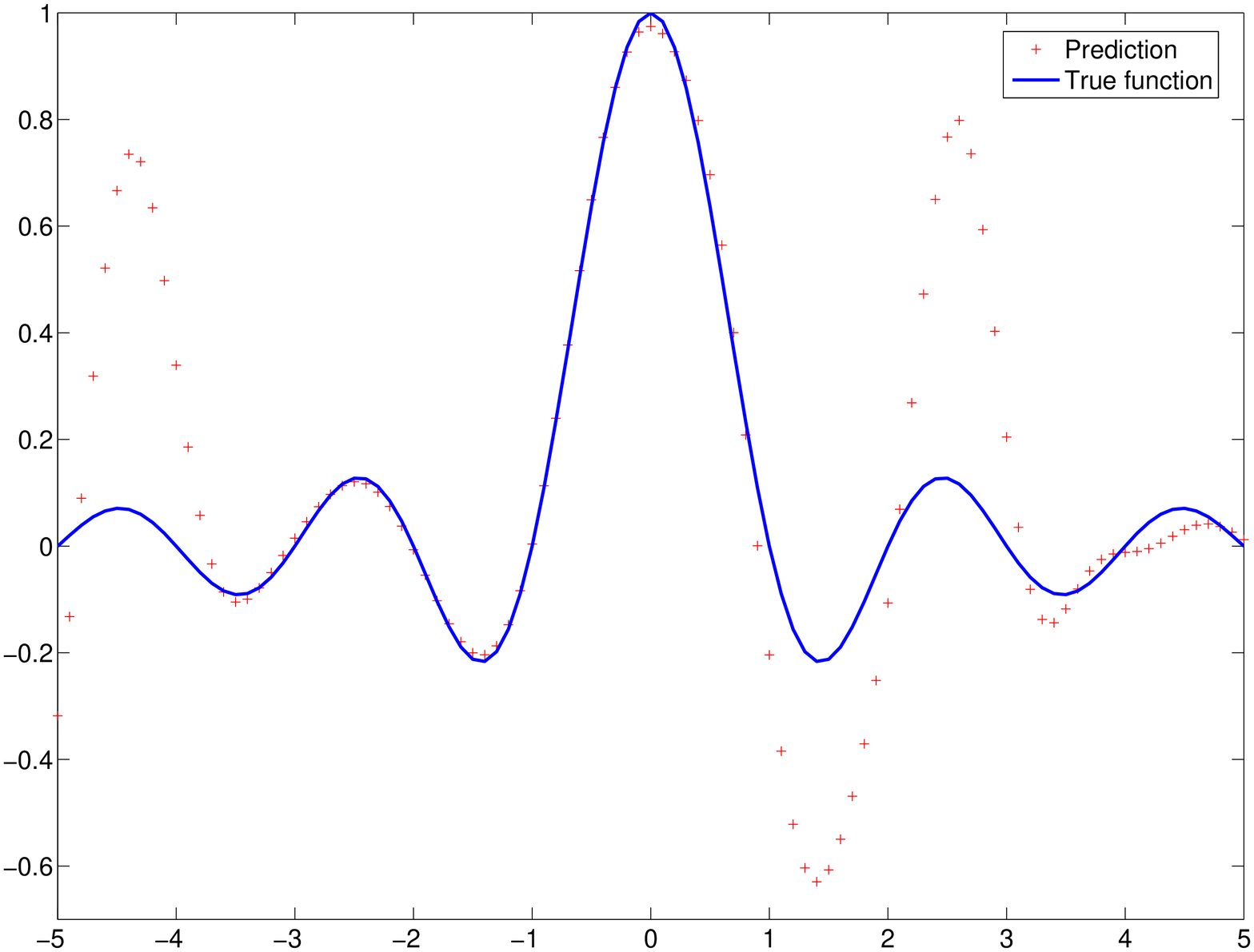}}
\vspace{-0.5cm}\centerline{(b)} 
\end{minipage}\\
\begin{minipage}[b]{0.4\linewidth}
  \centering
  \centerline{\includegraphics[width=\linewidth]{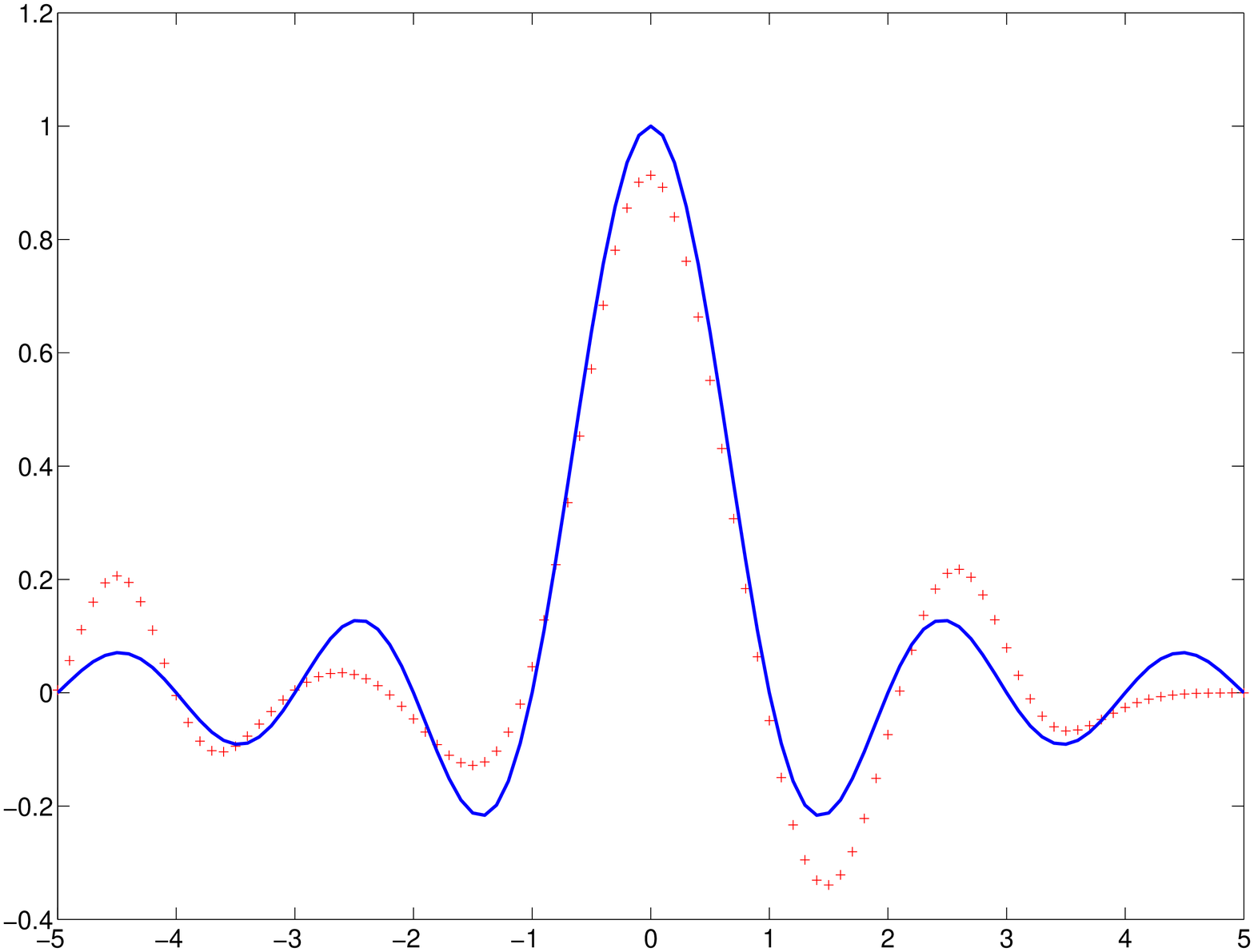}}
\vspace{-0.5cm}\centerline{(c)} 
\end{minipage}
\hfill
\begin{minipage}[b]{.4\linewidth}
  \centering
  \centerline{\includegraphics[width=\linewidth]{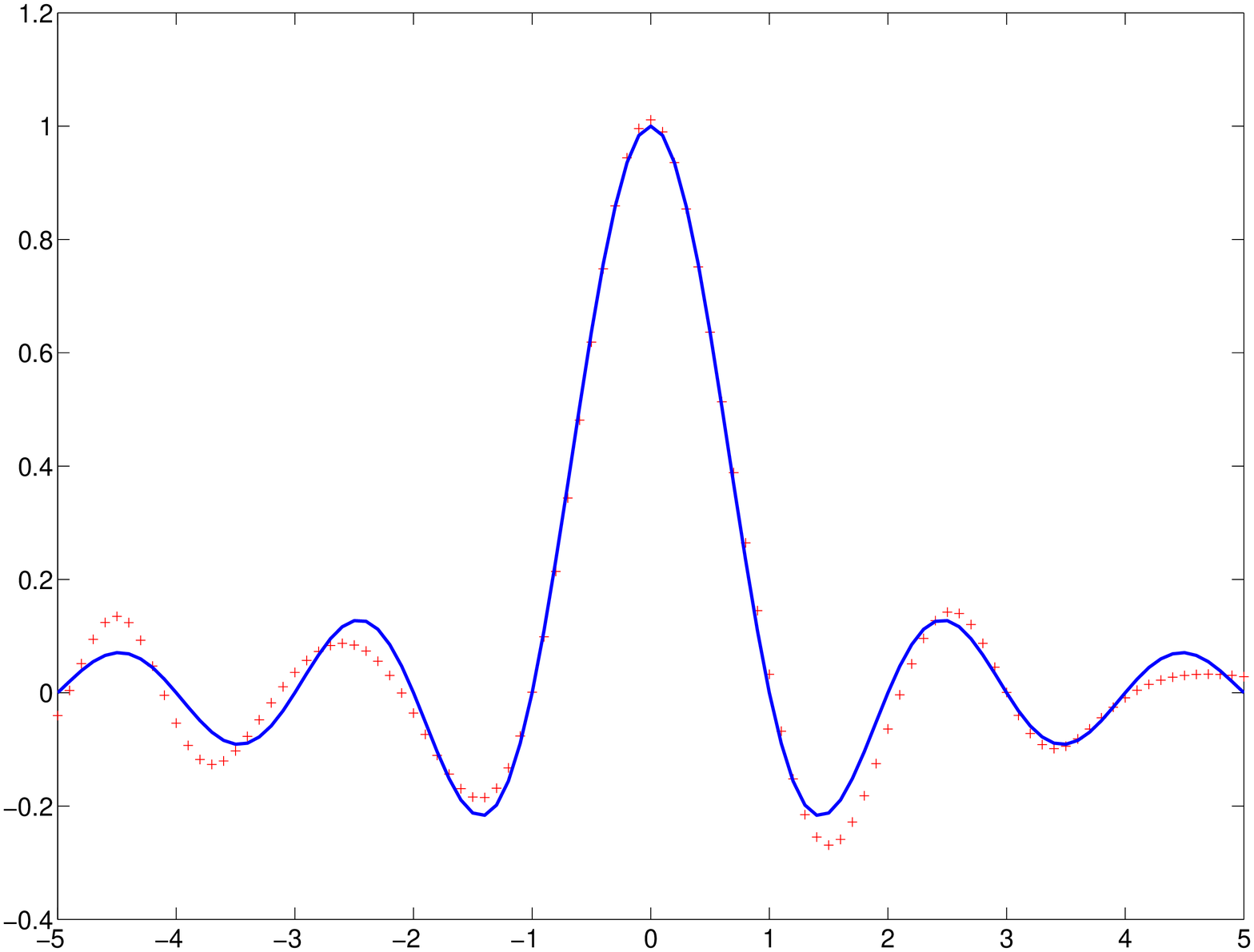}}
\vspace{-0.5cm}\centerline{(d)} 
\end{minipage}
\begin{minipage}[b]{0.4\linewidth}
  \centering
  \centerline{\includegraphics[width=\linewidth]{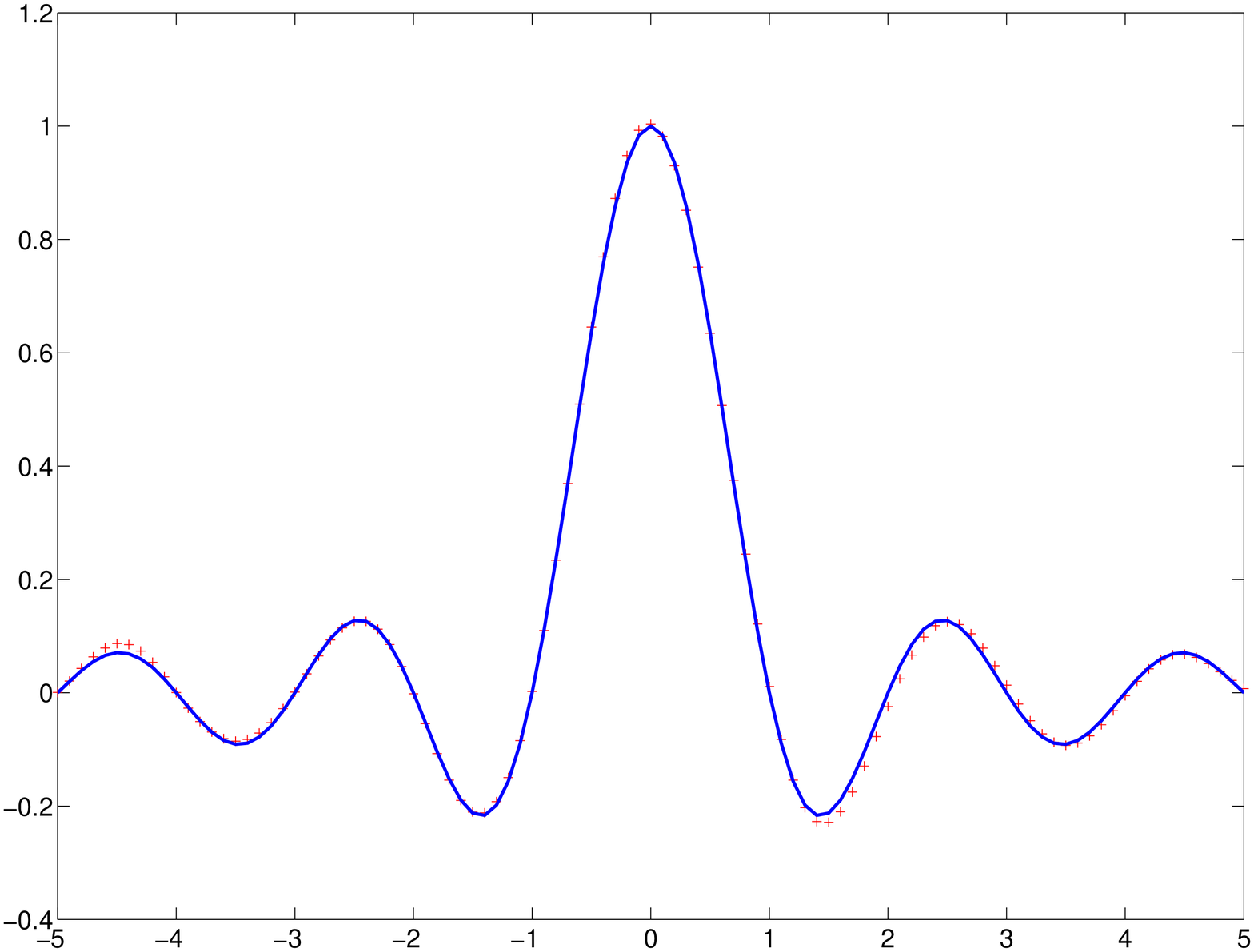}}
\vspace{-0.5cm}\centerline{(e)} 
\end{minipage}
\hfill
\begin{minipage}[b]{.4\linewidth}
  \centering
  \centerline{\includegraphics[width=\linewidth]{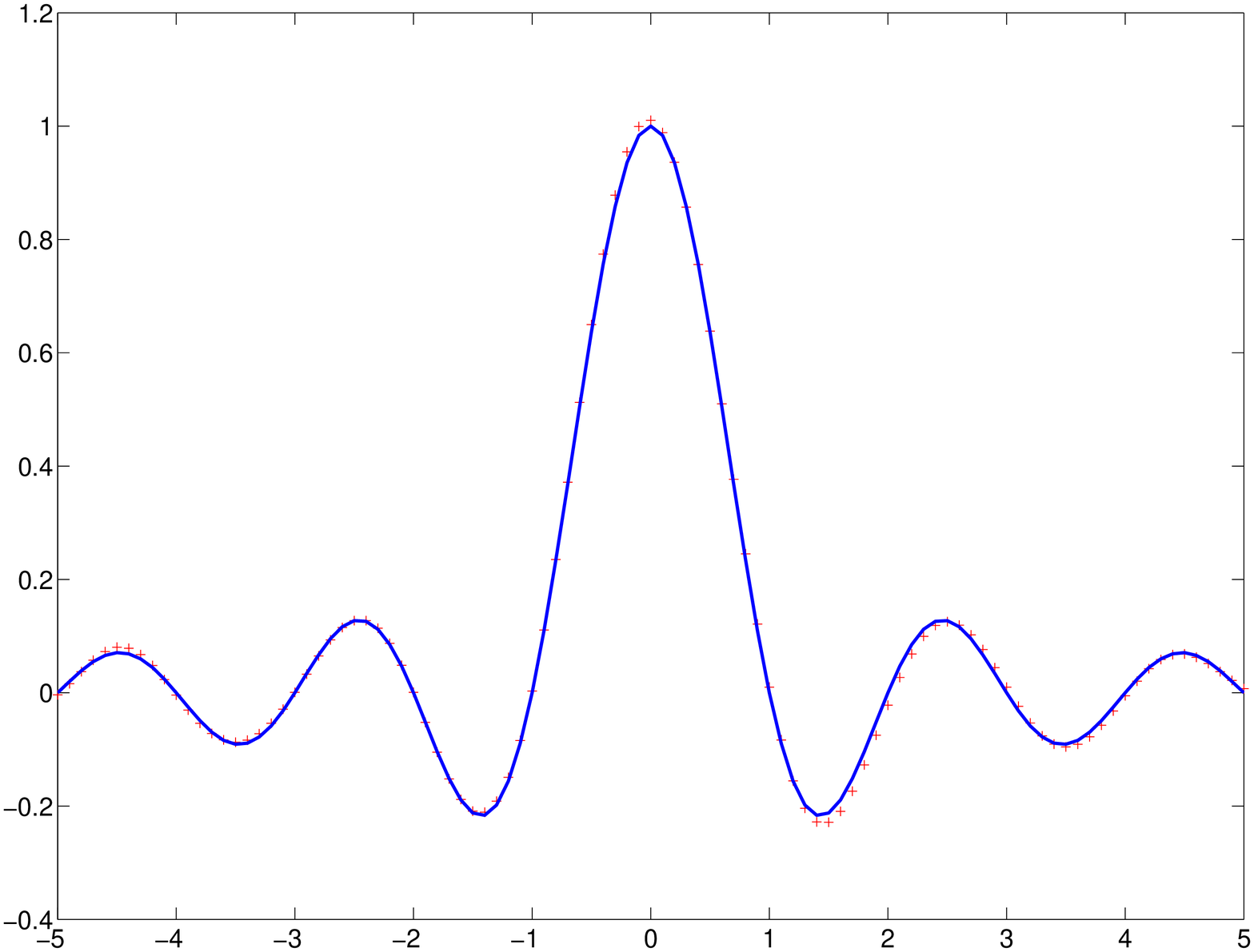}}
\vspace{-0.5cm}\centerline{(f)} 
\end{minipage}\\
\begin{minipage}[b]{0.4\linewidth}
  \centering
  \centerline{\includegraphics[width=\linewidth]{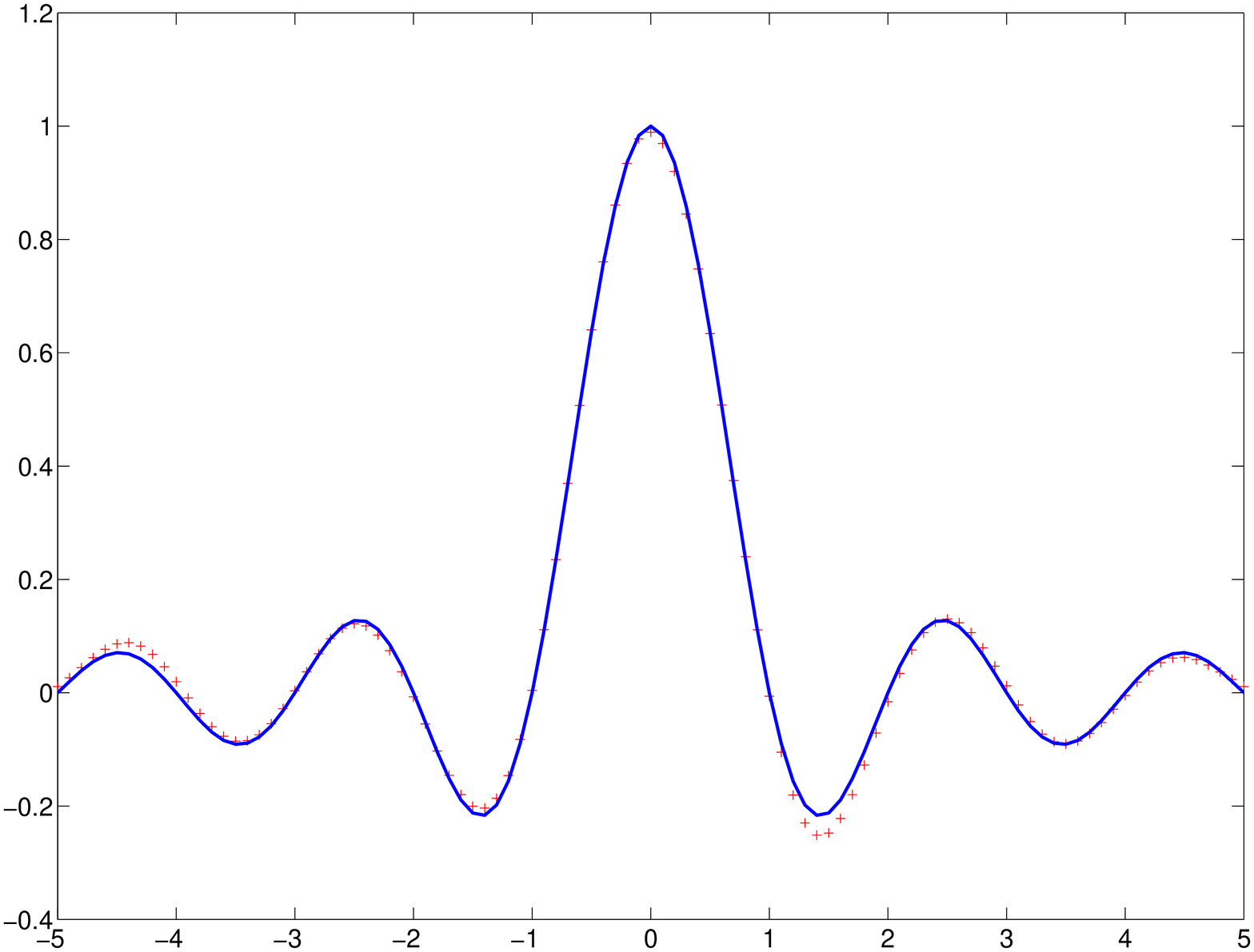}}
\vspace{-0.5cm}\centerline{(g)} 
\end{minipage}
\hfill
\begin{minipage}[b]{.4\linewidth}
  \centering
  \centerline{\includegraphics[width=\linewidth]{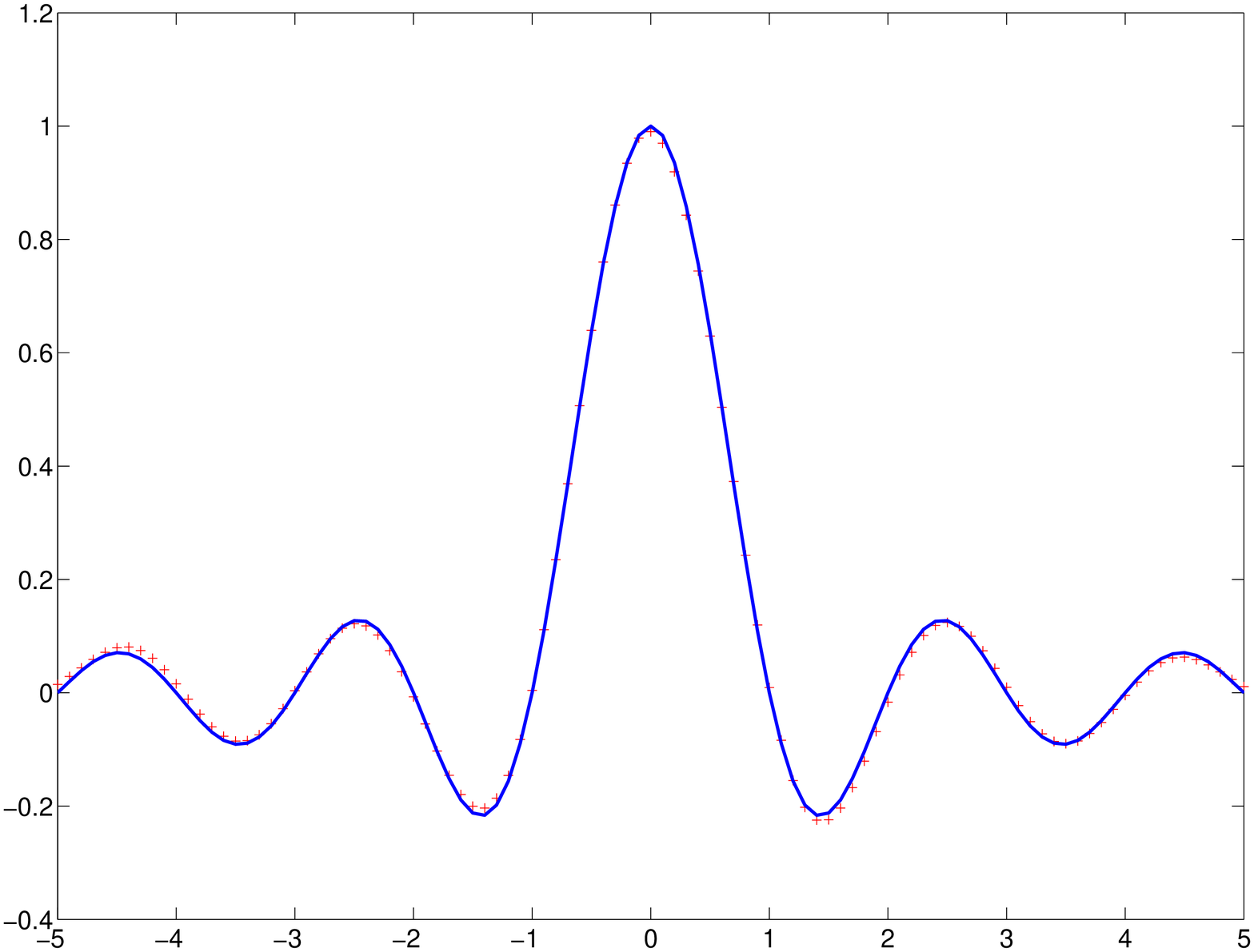}}
\vspace{-0.5cm}\centerline{(h)} 
\end{minipage}
\caption{Robust estimation of the $\textrm{sinc}$ function. The data is 
corrupted with $N_o=3$
outliers, and the nominal noise variance is $\sigma_{\varepsilon}^2=1\times 10^{-4}$. 
(a) Noisy training data and outliers;
(b) predicted values obtained after solving
\eqref{eq:regnets} with $V(u)=u^2$;
(c) SVR predictions for $\epsilon=0.1$; (d) RSVR predictions for 
$\epsilon=0.1$; 
(e) SVR predictions for $\epsilon=0.01$; (f) RSVR predictions for 
$\epsilon=0.01$; (g) predicted values obtained after 
solving
\eqref{eq:cost_l1unconstr}; (h) refined predictions using the
nonconvex regularization in \eqref{eq:cost_nonconvex}.}
\label{fig:Fig_4}
\end{figure}

The training set comprises $N=50$ examples,
with scalar inputs $\{x_i\}_{i=1}^N$ drawn from a uniform
distribution over $[-5,5]$. Uniformly distributed outliers 
$\{y_i\}_{i=1}^{N_o}\sim\mathcal{U}[-5,5]$ are artificially added in 
$\mathcal{T}$, with $N_o=3$ 
 resulting in $6\%$ contamination. Nominal data in $\mathcal{T}$ adheres to the
 model 
$y_i=\textrm{sinc}({x}_i)+\varepsilon_i$ for $i=N_o+1,\ldots,N$, where the
independent additive noise terms
$\varepsilon_i$ are zero-mean Gaussian
distributed. Three different values
are considered for the nominal noise variance, namely
$\sigma_{\varepsilon}^2=1\times 10^{-l}$ for $l=2,3,4$. For the case where
$\sigma_{\varepsilon}^2=1\times 10^{-4}$, the data used in the experiment are 
shown in Fig.
\ref{fig:Fig_4} (a). Superimposed to the true function $\textrm{sinc}(x)$ (shown in blue) 
are $47$ black points corresponding to the noisy data obeying the nominal 
model, as
 well as $3$ outliers depicted as red points.

The results are summarized in Table \ref{table:results_sinc}, 
which lists the 
generalization errors $\textrm{Err}_\mathcal{T}$ attained by the different 
methods tested, and for varying $\sigma_{\varepsilon}^2$. The independent test 
set 
$\{\tilde{y}_i,\tilde{x}_i\}_{i=1}^{\tilde{N}}$ used to evaluate 
\eqref{eq:gen_error} was generated from the model 
$\tilde{y}_i=\textrm{sinc}(\tilde{x}_i)+\varepsilon_i$, where the $\tilde{x}_i$
 define a  $\tilde{N}=101$-element uniform grid over $[-5,5]$. A first 
(expected) observation is that all robust alternatives markedly outperform the
 nonrobust regularization 
 network approach in \eqref{eq:regnets}, by an order of magnitude or even more,
 regardless of the value of $\sigma_{\varepsilon}^2$. As reported 
in~\cite{RSVR}, RSVR uniformly outperforms SVR. For the case 
$\epsilon=0.01$, RSVR also uniformly outperforms the sparsity-controlling 
method in \eqref{eq:cost_l1unconstr}. Interestingly, after refining the 
estimate obtained via \eqref{eq:cost_l1unconstr} through a couple iterations of
 \eqref{eq:iters_nonconvex_o} (cf. Section \ref{sec:nonconvex}), the lowest 
generalization errors are obtained, uniformly across all simulated 
values of the nominal noise variance. Results for the RSVR with $\epsilon=0.01$
 come sufficiently close, and are equally satisfactory for all practical 
purposes; see also Fig. \ref{fig:Fig_4} for a pictorial summary of the
 results when $\sigma_{\varepsilon}^2=1\times 10^{-4}$.

While specific error values or method rankings are arguably 
anecdotal, 
two conclusions stand out: (i) model \eqref{eq:model_outliers} and 
its sparsity-controlling estimators \eqref{eq:cost_l1unconstr} and 
\eqref{eq:cost_nonconvex} are effective approaches to nonparametric 
regression in the presence of outliers; and (ii) when initialized with 
$\hat{\mathbf{o}}_{\textrm{Lasso}}$ the refined estimator 
\eqref{eq:cost_nonconvex} can 
considerably improve the performance of \eqref{eq:cost_l1unconstr}, at the 
price of a modest increase in computational complexity. While 
\eqref{eq:cost_l1unconstr} endowed with the sparsity-controlling mechanisms 
of Section \ref{ssec:params} tends to overestimate the ``true'' support of 
$\mathbf{o}$, numerical results have consistently shown that the refinement in 
Section \ref{sec:nonconvex} is more effective when it comes to support 
recovery.

\subsection{Load curve data cleansing}\label{ssec:real_data}
In this section, the robust nonparametric methods described so far are applied 
to the 
problem of load curve cleansing outlined in Section \ref{sec:intro}. Given 
load data $\mathcal{T}:=\{y_i,t_i\}_{i=1}^{N}$ corresponding to a building's 
power 
consumption 
measurements $y_i$, acquired at time instants $t_i$, $i=1,\ldots,N$, the 
proposed approach to load curve cleansing minimizes
\begin{equation}\label{eq:robust_smoothing_splines}
\min_{\substack{f\in\mathcal{S}\\\mathbf{o}\in\mathbb{R}^N}}
\left[\sum_{i=1}^{N}(y_{i}-f(t_i)-o_i)^2
+\mu\int_{\mathbb{R}}f''(t) dt+\lambda_1\|\mathbf{o}\|_1\right]
\end{equation}
where $f''(t)$ denotes the second-order derivative of 
$f:\mathbb{R}\to\mathbb{R}$. 
This way, the solution $\hat f$ provides a cleansed estimate of the load 
profile, and the
 support of $\hat{\mathbf{o}}$ indicates the instants where significant load 
deviations, or, meter failures occurred. 
Estimator \eqref{eq:robust_smoothing_splines} specializes 
\eqref{eq:cost_l1unconstr} to the so-termed \textit{cubic smoothing splines}; 
see e.g.,~\cite{elements_of_statistics,wahba}. It is also subsumed as a special
 case of the robust thin-plate splines estimator \eqref{eq:thin_plate_splines},
 when 
the target function $f$ has domain in $\mathbb{R}$ [cf. how the smoothing 
penalty \eqref{eq:smoothing_penalty} simplifies to the one in 
\eqref{eq:robust_smoothing_splines} in the one-dimensional case].

In light of the aforementioned connection, it should not be 
surprising that 
$\hat{f}$ admits a unique, finite-dimensional minimizer, which corresponds to a
 \textit{natural spline} with knots at $\{t_i\}_{i=1}^N$; see 
e.g.,~\cite[p. 151]{elements_of_statistics}. Specifically, it follows that 
$\hat f(t)=\sum_{i=1}^N\hat{\theta}_i b_i(t)$, where $\{b_i(t)\}_{i=1}^N$ is 
the
 basis set of natural spline functions, and the vector of expansion 
coefficients $\hat{\bm \theta}:=[\hat\theta_1,\ldots,\hat\theta_N]^\prime$ is 
given by
\begin{equation*}\label{eq:theta_sol}
\hat{\bm\theta}=\left(\mathbf{B}^{\prime}\mathbf{B}+\mu\bm\Psi\right)^{-1}
\mathbf{B}^{\prime}(\mathbf{y}-\hat{\mathbf{o}})
\end{equation*}
where matrix $\mathbf{B}\in\mathbb{R}^{N\times N}$ has $ij$-th entry 
$[\mathbf{B}]_{ij}=b_j(t_i)$; while $\bm\Psi\in\mathbb{R}^{N\times N}$ has 
$ij$-th entry $[\bm\Psi]_{ij}=\int b_i''(t)b_j''(t) dt$. Spline coefficients 
can be computed more efficiently if the basis of B-splines is adopted instead; 
details can be found in~\cite[p. 189]{elements_of_statistics} 
and~\cite{unser_tutorial}.

Without 
considering the outlier variables in \eqref{eq:robust_smoothing_splines}, a 
B-spline estimator for load curve cleansing was put forth
in~\cite{load_curve_cleansing}. An alternative Nadaraya-Watson estimator from 
the Kernel smoothing family was considered as well. In any case, outliers are 
identified 
during a 
post-processing 
stage, after the load curve has been estimated nonrobustly. Supposing for 
instance that the approach 
in~\cite{load_curve_cleansing} correctly identifies outliers most of the time, 
it still does not yield a cleansed estimate $\hat f$. This should be contrasted
 with the estimator \eqref{eq:robust_smoothing_splines}, which accounts for the
 outlier compensated data to yield a cleansed estimate at once. Moreover, to 
select the ``optimum'' smoothing parameter $\mu$, the approach 
of~\cite{load_curve_cleansing} requires the user to manually label
  the outliers present in a training subset of data, during a pre-processing 
stage. This subjective component makes it challenging to reproduce the results 
of~\cite{load_curve_cleansing}, and for this reason comparisons with the 
aforementioned scheme are not included in the sequel.

\begin{figure}[h]
\centering
\begin{minipage}[b]{0.8\linewidth}
  \centering
  \centerline{\includegraphics[width=\linewidth]{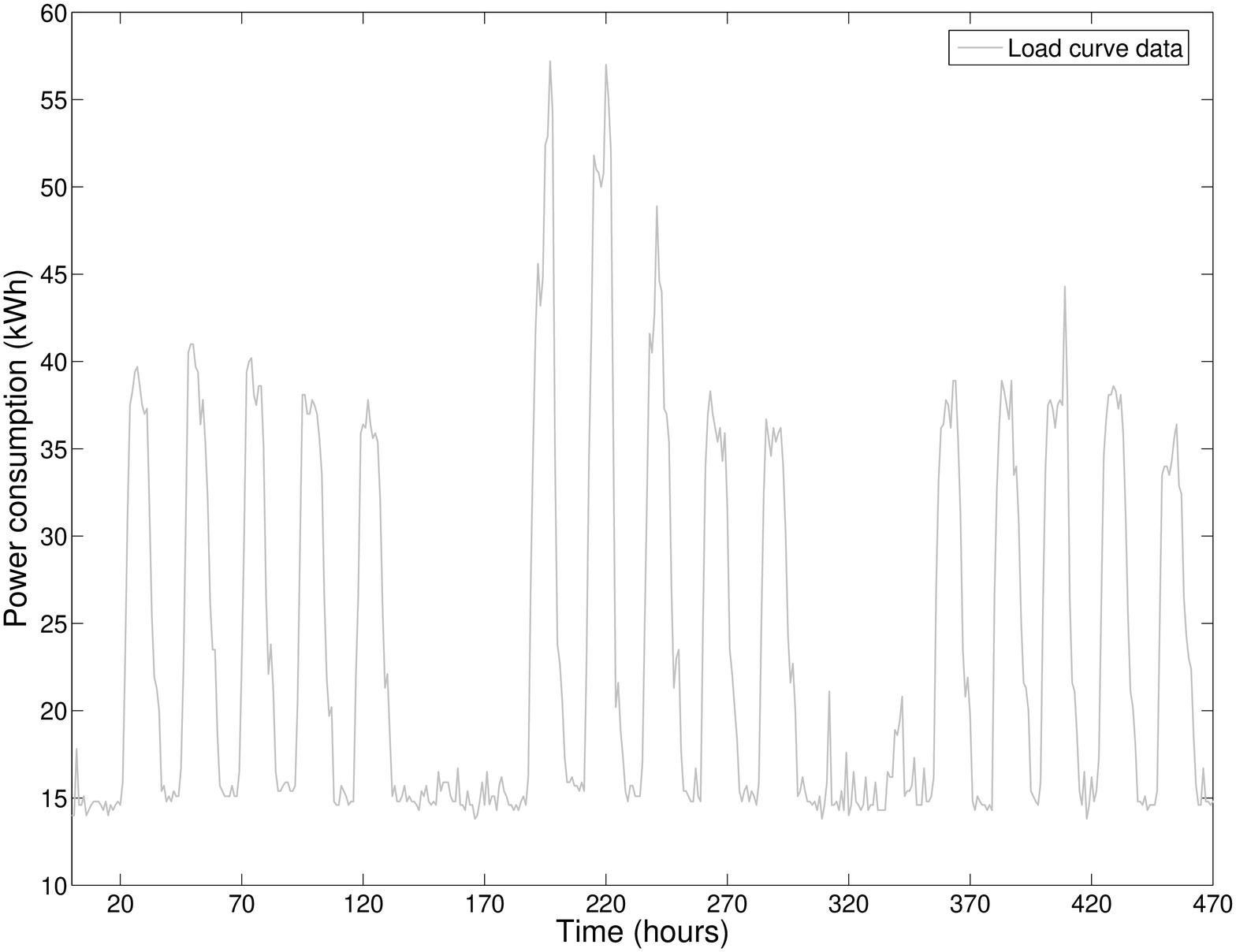}}
\vspace{-0.5cm}\centerline{(a)} 
\end{minipage}\\
\begin{minipage}[b]{0.8\linewidth}
  \centering
  \centerline{\includegraphics[width=\linewidth]{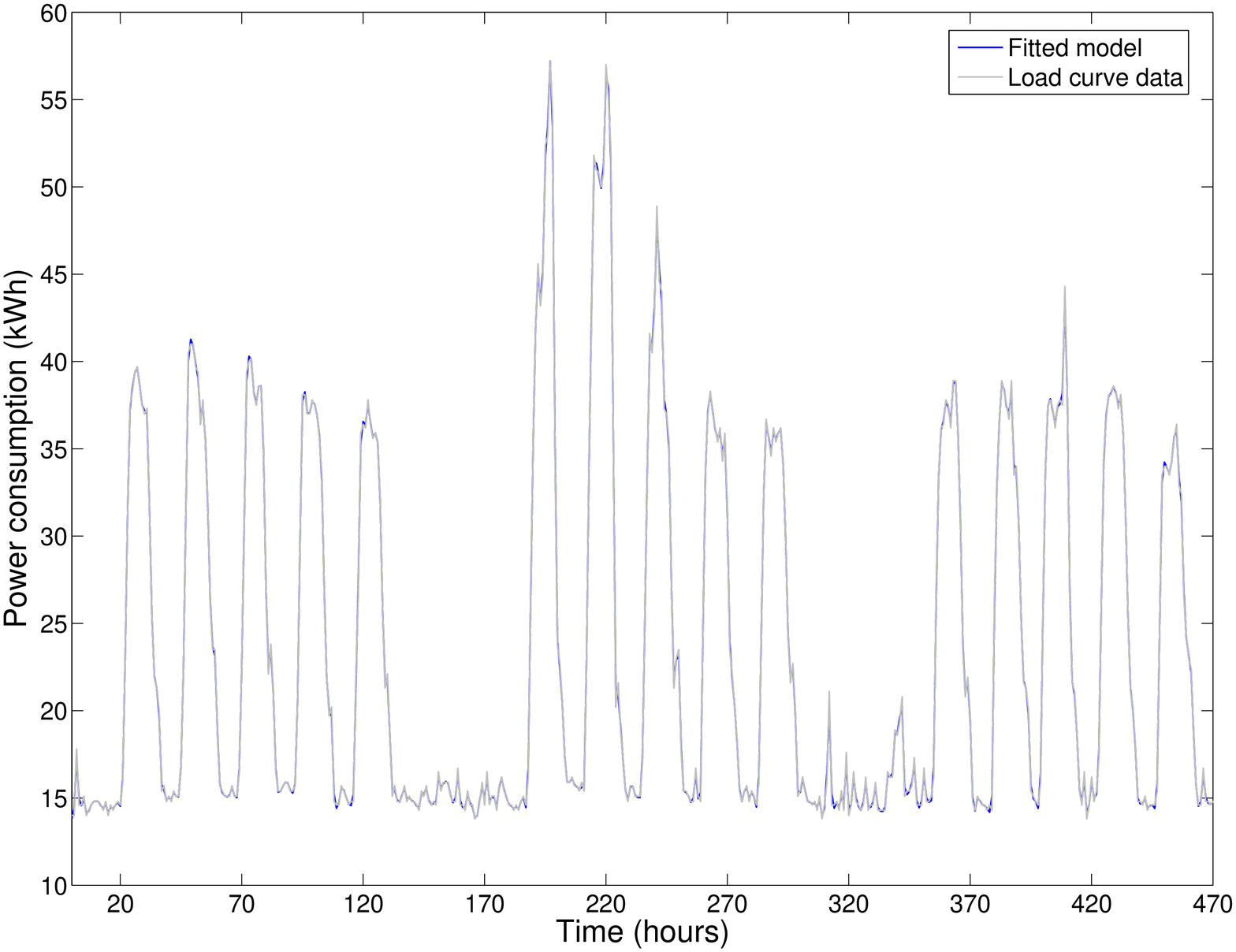}}
\vspace{-0.5cm}\centerline{(b)} 
\end{minipage}
\caption{Load curve data cleansing. 
(a) Noisy training data and outliers;
(b) fitted load profile obtained after solving
\eqref{eq:nonrobust_smoothing_splines}.}
\label{fig:Fig_5}
\end{figure}

Next, estimator \eqref{eq:robust_smoothing_splines} is tested 
on real load curve data provided by the NorthWrite
Energy Group. The dataset consists of power consumption measurements (in kWh) 
for a government building, collected every fifteen minutes during a period of 
more than five years, ranging from July 2005 to October 2010. Data is 
downsampled by a 
factor of four, to yield one measurement per hour. For the present experiment, 
only a subset of the whole data is utilized for concreteness, where $N=501$ 
was chosen corresponding to a $501$ hour period. A snapshot of this training
 load curve data in $\mathcal{T}$, spanning a particular
three-week period is shown in Fig. \ref{fig:Fig_5} (a). Weekday activity 
patterns 
can be clearly discerned from those corresponding to weekends, as expected for
 most government buildings; but different, e.g., for the load profile of a 
grocery store. Fig. \ref{fig:Fig_5} (b) shows the nonrobust smoothing spline 
fit to the training data in $\mathcal{T}$ (also shown for comparison 
purposes), obtained after solving
\begin{equation}\label{eq:nonrobust_smoothing_splines}
\min_{f\in\mathcal{S}}
\left[\sum_{i=1}^{N}(y_{i}-f(t_i))^2
+\mu\int_{\mathbb{R}}f''(t) dt\right]
\end{equation}
using Matlab's built-in spline toolbox. Parameter $\mu$ was chosen based on 
leave-one-out cross-validation, and it is apparent that no cleansing of the 
load profile takes place. Indeed, the resulting fitted function follows very 
closely the training data, even during the abnormal energy peaks observed 
on the so-termed ``building operational transition shoulder periods.''

\begin{figure}[h]
\centering
\begin{minipage}[b]{0.8\linewidth}
  \centering
  \centerline{\includegraphics[width=\linewidth]{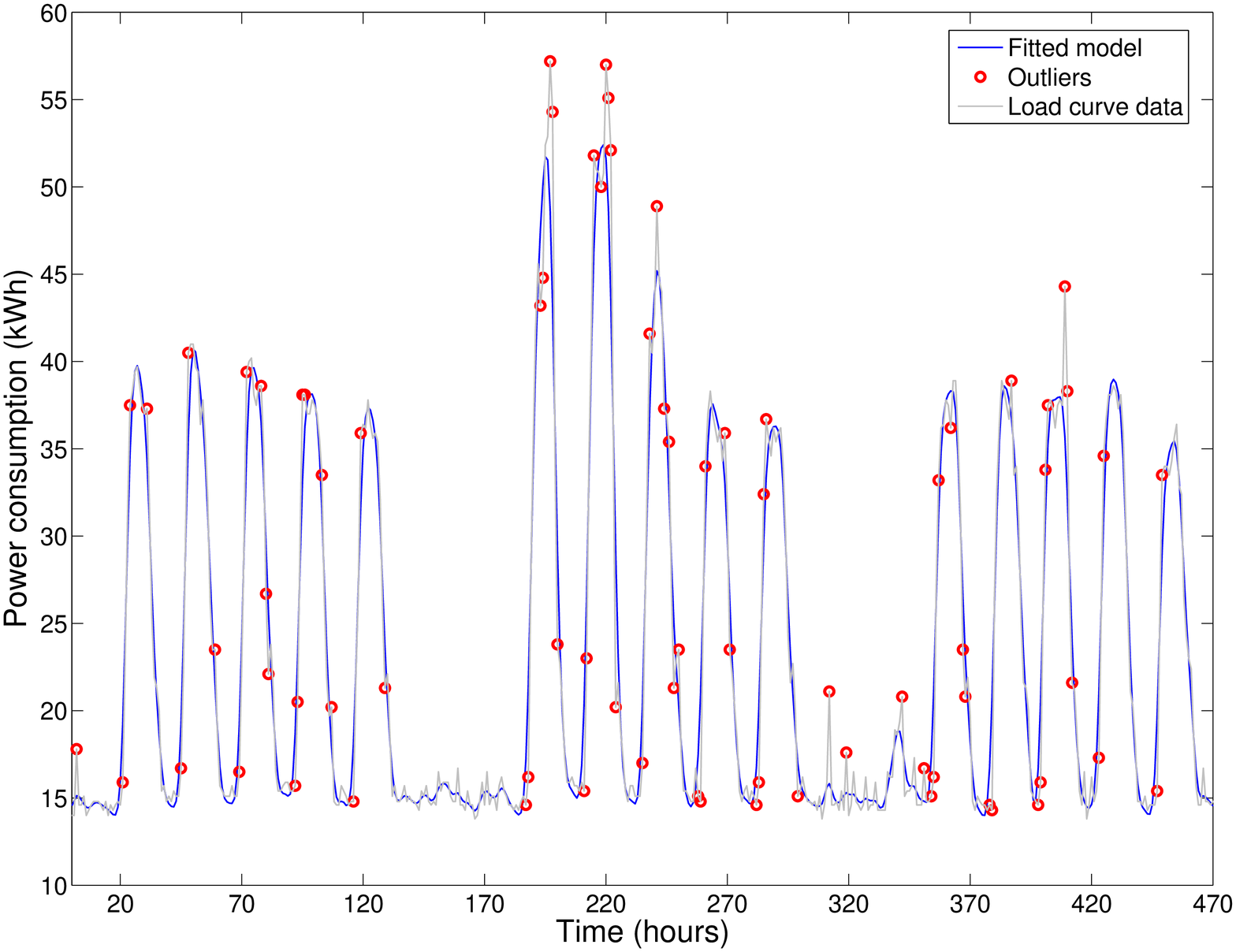}}
\vspace{-0.5cm}\centerline{(a)} 
\end{minipage}\\
\begin{minipage}[b]{0.8\linewidth}
  \centering
  \centerline{\includegraphics[width=\linewidth]{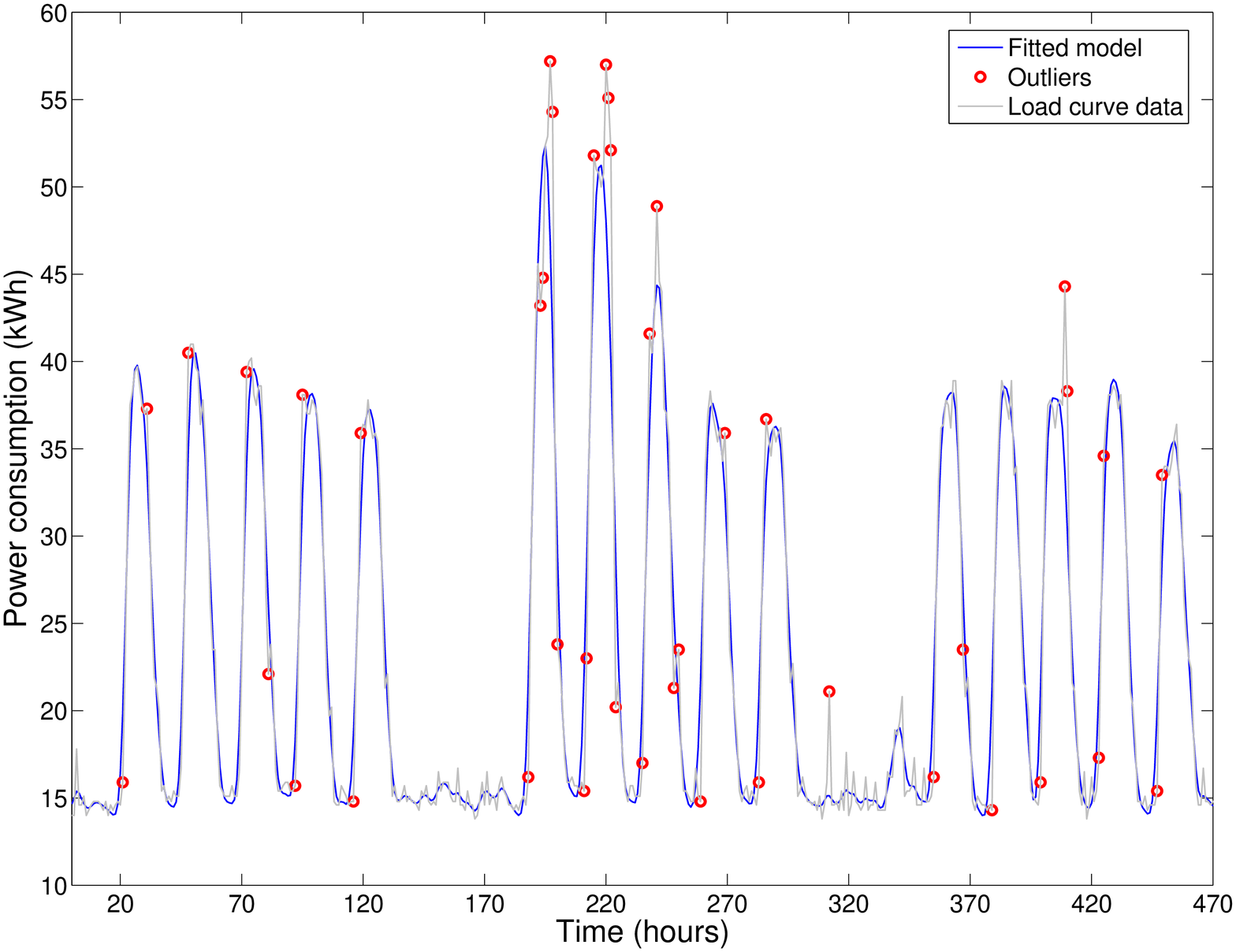}}
\vspace{-0.5cm}\centerline{(b)} 
\end{minipage}
\caption{Load curve data cleansing. 
(a) Cleansed load profile obtained after solving 
\eqref{eq:robust_smoothing_splines};
(b) refined  load profile obtained after using the nonconvex regularization in 
\eqref{eq:cost_nonconvex}.}
\label{fig:Fig_6}
\end{figure}

Because with real load curve data the nominal noise variance 
$\sigma_{\varepsilon}^2$ in \eqref{eq:model_outliers} is unknown, selection of 
the tuning parameters $\{\mu,\lambda_1\}$ in 
\eqref{eq:robust_smoothing_splines} requires a robust estimate of the 
variance $\hat{\sigma}_{\varepsilon}^2$ such as the MAD [cf. Section 
\ref{ssec:params}]. Similar to~\cite{load_curve_cleansing}, it is assumed that 
the nominal errors are zero mean Gaussian distributed, so that \eqref{eq:MAD} 
can be applied yielding the value $\hat{\sigma}_{\varepsilon}^2=0.6964$. To 
form the
 residuals in \eqref{eq:MAD}, \eqref{eq:nonrobust_smoothing_splines} is solved 
first using a small subset of $\mathcal{T}$ that comprises $126$ measurements. 
 A nonuniform
grid of $\mu$ and $\lambda_1$ values is constructed, as
described in Section \ref{ssec:params}. Relevant parameters
are $G_{\mu}=100$, $G_{\lambda}=200$, $\mu_{\min}=10^{-3}$,
$\mu_{\max}=10$, and
$\epsilon=10^{-4}$. The robustification paths (one per $\mu$ value in the grid)
were obtained
using the SpaRSA toolbox in~\cite{Wright_tsp_2009}, with the sample variance 
matrix
$\bar{\bm\Sigma}$ formed as in \eqref{eq:sample_variance}.  The optimum tuning 
parameters $\mu^{\ast}=1.637$ and $\lambda_1^{\ast}=3.6841$ are finally 
determined
based on the criterion \eqref{eq:select_ij}, where the unknown 
$\sigma_{\varepsilon}^2$ is replaced with $\hat{\sigma}_{\varepsilon}^2$. 
Finally, the
cleansed load curve is refined by running four iterations of
\eqref{eq:iters_nonconvex_o} as described in Section
\ref{sec:nonconvex}, with a value of $\delta=10^{-5}$. Results are depicted in 
Fig. \ref{fig:Fig_6}, where the cleansed load curves are superimposed to 
the training data in $\mathcal{T}$. Red circles indicate those data points 
deemed 
as outliers, information that is readily obtained from the support of 
$\hat{\mathbf{o}}$. By inspection of Fig. \ref{fig:Fig_6}, it is apparent that 
the proposed sparsity-controlling estimator has the desired cleansing 
capability.
 The cleansed load curves closely follow the training data, but are smooth 
enough 
to avoid overfitting the abnormal energy peaks on the ``shoulders.'' Indeed, 
these peaks are in most cases identified as outliers. As seen from Fig.
 \ref{fig:Fig_6} (a), the solution of \eqref{eq:robust_smoothing_splines} 
tends to overestimate the support of $\mathbf{o}$, since one could argue that 
some of the red circles in Fig. \ref{fig:Fig_6} (a) do not correspond to 
outliers. 
Again, the 
nonconvex regularization in Section \ref{sec:nonconvex} prunes the outlier 
support obtained
 via \eqref{eq:robust_smoothing_splines}, resulting in a more accurate result 
and
 reducing the number of outliers identified from $77$ to $41$.

\section{Concluding Summary}\label{sec:conclusion}

Outlier-robust nonparametric regression methods were developed 
in this paper for function approximation in RKHS. Building on a neat link 
between the seemingly unrelated fields of robust statistics and sparse  
regression, the novel estimators were found rooted at the crossroads of 
outlier-resilient estimation, the Lasso, and convex optimization. Estimators as
 fundamental as LS for linear regression, regularization networks, and 
(thin-plate) smoothing splines, can be robustified under the proposed 
framework.

Training samples from the (unknown) target function were 
assumed generated from a regression 
model, which explicitly incorporates an unknown sparse vector of outliers. To 
fit such a model, the proposed variational estimator minimizes a
 tradeoff 
between fidelity to the training data, the degree of ``smoothness'' of the 
regression function, and the sparsity level of the vector of outliers. While 
model complexity control effected through a smoothing penalty has quite well 
understood ramifications in terms of generalization capability,
 the major innovative claim here is that \textit{sparsity control} is 
tantamount to robustness control. This is indeed the case since a tunable 
parameter in a Lasso reformulation of the variational estimator, controls the 
degree of sparsity in the estimated vector 
of model outliers. 
Selection of tuning parameters could be at first thought as a mundane task. 
However, arguing on the importance of such task in the context of robust 
nonparametric regression, as well as devising principled
 methods to effectively carry out smoothness and sparsity control, are
 at the heart of this paper's novelty. Sparsity control can be carried out at 
affordable complexity, by capitalizing on state-of-the-art algorithms that can 
efficiently compute the whole path of Lasso solutions. In this sense, the 
method 
here capitalizes on but is not limited to sparse settings where few outliers 
are present, since one can efficiently examine the gamut of sparsity levels 
along the robustification path. Computer simulations have shown that the novel 
methods of this paper outperform existing alternatives including SVR, and one 
if its robust variants.

As an application domain relevant to robust nonparametric 
regression, the problem of load curve cleansing for power systems engineering 
was also considered along with a 
solution proposed based on robust cubic spline smoothing. Numerical tests on 
real load curve data demonstrated that the smoothness and sparsity controlling 
methods of this paper are effective in cleansing load profiles, without 
user intervention to aid the learning process.

\section*{Acknowledgment}  The authors would like to thank NorthWrite
Energy Group and Prof. Vladimir
Cherkassky (Dept. of ECE, University of Minnesota) for providing
the load curve data analysed in Section \ref{ssec:real_data}.

{\Large\appendix}

Towards establishing the equivalence between problems 
\eqref{eq:cost_l1unconstr}
and \eqref{eq:variational_w_rho},
consider the pair $\{\hat f,\hat{\mathbf{o}}\}$ that solves
\eqref{eq:cost_l1unconstr}. Assume that $\hat f$ is given, and
the goal is to determine $\hat{\mathbf{o}}$.
Upon defining the residuals $\hat r_i:=y_i-\hat
f(\mathbf{x}_i)$ and because
$\|\mathbf{o}\|_1=\sum_{i=1}^N|o_i|$, the
entries of $\hat{\mathbf{o}}$ are separately given by
\begin{equation}\label{eq:cost_oi_convex}
\hat{o}_i:=\arg\min_{o_i\in\mathbb{R}}
\left[(\hat{r}_{i}-o_i)^2+\lambda_1|o_i|\right],\quad
i=1,\ldots,N,
\end{equation}
where the term $\mu\|\hat f\|_{\mathcal{H}}^2$ in
\eqref{eq:cost_l1unconstr} has been omitted, since it is
inconsequential for the minimization with respect to
$\mathbf{o}$. For each $i=1,\ldots,N$, because \eqref{eq:cost_oi_convex}
is nondifferentiable at the origin one should consider three
cases: i) if $\hat{o}_i=0$, it follows that the minimum cost in
\eqref{eq:cost_oi_convex} is $\hat{r}_{i}^2$; ii) if
$\hat{o}_i> 0$, the first-order condition for optimality
gives $\hat{o}_i=\hat{r}_{i}-\lambda_1/2$ provided 
$\hat{r}_{i}>\lambda_1/2$, and the minimum cost is 
$\lambda_1\hat{r}_{i}-\lambda_1^2/4$; otherwise, iii) if
$\hat{o}_i< 0$, it follows that $\hat{o}_i=\hat{r}_{i}+\lambda_1/2$ provided 
$\hat{r}_{i}<-\lambda_1/2$, and the minimum cost is 
$-\lambda_1\hat{r}_{i}-\lambda_1^2/4$. In other words,
\begin{equation}\label{eq:min_oi_2}
\hat{o}_i=\left\{\begin{array}{ccc}
\hat{r}_{i}-\lambda_1/2,&&\hat{r}_{i}>\lambda_1/2\\
0,&&|\hat{r}_{i}|\leq\lambda_1/2\\
\hat{r}_{i}+\lambda_1/2,&&\hat{r}_{i}<-\lambda_1/2\end{array}\right.,\quad i=1,\ldots,N.
\end{equation}
Upon plugging \eqref{eq:min_oi_2} into \eqref{eq:cost_oi_convex}, the
minimum cost in \eqref{eq:cost_oi_convex} after minimizing with
respect to $o_i$ is $\rho(\hat r_i)$ [cf. \eqref{eq:rho_def} and
the argument preceding \eqref{eq:min_oi_2}].
All in all, the conclusion is that $\hat f$
is the minimizer of 
\eqref{eq:variational_w_rho} -- in addition to being the solution
of \eqref{eq:cost_l1unconstr} by definition -- completing the proof.
\hfill$\blacksquare$

\bibliographystyle{IEEEtranS}
\bibliography{IEEEabrv,tsp-rnr}

\newpage

%

\end{document}